%% file: main.tex
\documentclass[10pt,twocolumn,letterpaper]{article}

\usepackage[pagenumbers]{cvpr}   %

\input{preamble}

\usepackage{booktabs}
\usepackage{multirow}
\usepackage{amsmath,amssymb}
\usepackage{graphicx}
\usepackage{tikz}
\usetikzlibrary{positioning}
\usepackage{fontawesome5}
\usepackage{twemojis}

\newcommand{\stageone}{Stage~1}
\newcommand{\stagetwo}{Stage~2}
\newcommand{\ours}{ALPS}

\usepackage{colortbl}
\definecolor{rk1}{RGB}{107,174,214} %
\definecolor{rk2}{RGB}{158,202,225}
\definecolor{rk3}{RGB}{198,219,239}
\definecolor{oursrow}{RGB}{255,247,232}  %

\definecolor{cvprblue}{rgb}{0.21,0.49,0.74}
\usepackage{xr-hyper}
\usepackage[pagebackref,breaklinks,colorlinks,allcolors=cvprblue]{hyperref}
\ifdefined\NOSUPP\fi
\ifdefined\SUPPONLY\fi
\newsavebox{\fitbox}
\newcommand{\fitwidth}[2]{\sbox{\fitbox}{#2}\ifdim\wd\fitbox>#1 \resizebox{#1}{!}{\usebox{\fitbox}}\else\usebox{\fitbox}\fi}
\def\paperID{*****}
\def\confName{CVPR}
\def\confYear{2027}

\title{\raisebox{-0.07em}{\includegraphics[height=0.82em]{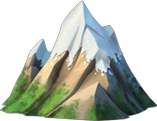}}\hspace{0.18em}\ours{}: Visual State Encoders for World Models}

\author{
Yanqing Liu$^{1}$ \quad Letian Zhang$^{1}$ \quad Chengzhi Liu$^{3}$ \quad Zijun Wang$^{1}$\\[5pt]
Jieru Mei$^{2}$ \quad Hongliang Fei$^{2}$ \quad Yuyin Zhou$^{1}$ \quad Cihang Xie$^{1}$\\[8pt]
$^{1}$University of California, Santa Cruz \quad $^{2}$Google \quad $^{3}$University of California, Santa Barbara\\[11pt]
{\small\hypersetup{urlcolor=black!80}\color{black!80}%
\faGlobe\enspace\href{https://ucsc-vlaa.github.io/ALPS}{\texttt{ucsc-vlaa.github.io/ALPS}}\hspace{1em}%
\faGithub\enspace\href{https://github.com/UCSC-VLAA/ALPS}{\texttt{github.com/UCSC-VLAA/ALPS}}\hspace{1em}%
\raisebox{-0.15em}{\twemoji[height=1em]{hugs}}\enspace\href{https://huggingface.co/UCSC-VLAA/ALPS}{\texttt{huggingface.co/UCSC-VLAA/ALPS}}}
}
\makeatletter
\renewcommand\paragraph{\@startsection{paragraph}{4}{\z@}{0pt plus 1ex}{-1em}{\normalfont\normalsize\bfseries}}
\makeatother

\begin{document}
\ifdefined\SUPPONLY\else
\maketitle
\input{fig/teaser}
\input{sec/0_abstract}
\input{sec/1_intro}
\input{sec/2_related}
\input{sec/3_method}
\input{sec/4_experiments}
\input{sec/5_analysis}
\input{sec/6_conclusion}
{
    \small
    \bibliographystyle{ieeenat_fullname}
    \bibliography{main}
}
\fi

\ifdefined\NOSUPP\else
\ifdefined\SUPPONLY\else\clearpage\fi
\input{sec/X_suppl}
\ifdefined\SUPPONLY
{
    \small
    \bibliographystyle{ieeenat_fullname}
    \bibliography{main}
}
\fi
\fi

\end{document}

%% file: preamble.tex
\usepackage{cuted} %

\usepackage{currfile} %

\usepackage{caption} %

%% file: fig/teaser.tex
\preCutedStrip{\vskip-12pt}
\begin{strip}
\centering
\includegraphics[width=1\textwidth,trim=0 5.5 0 4.2,clip]{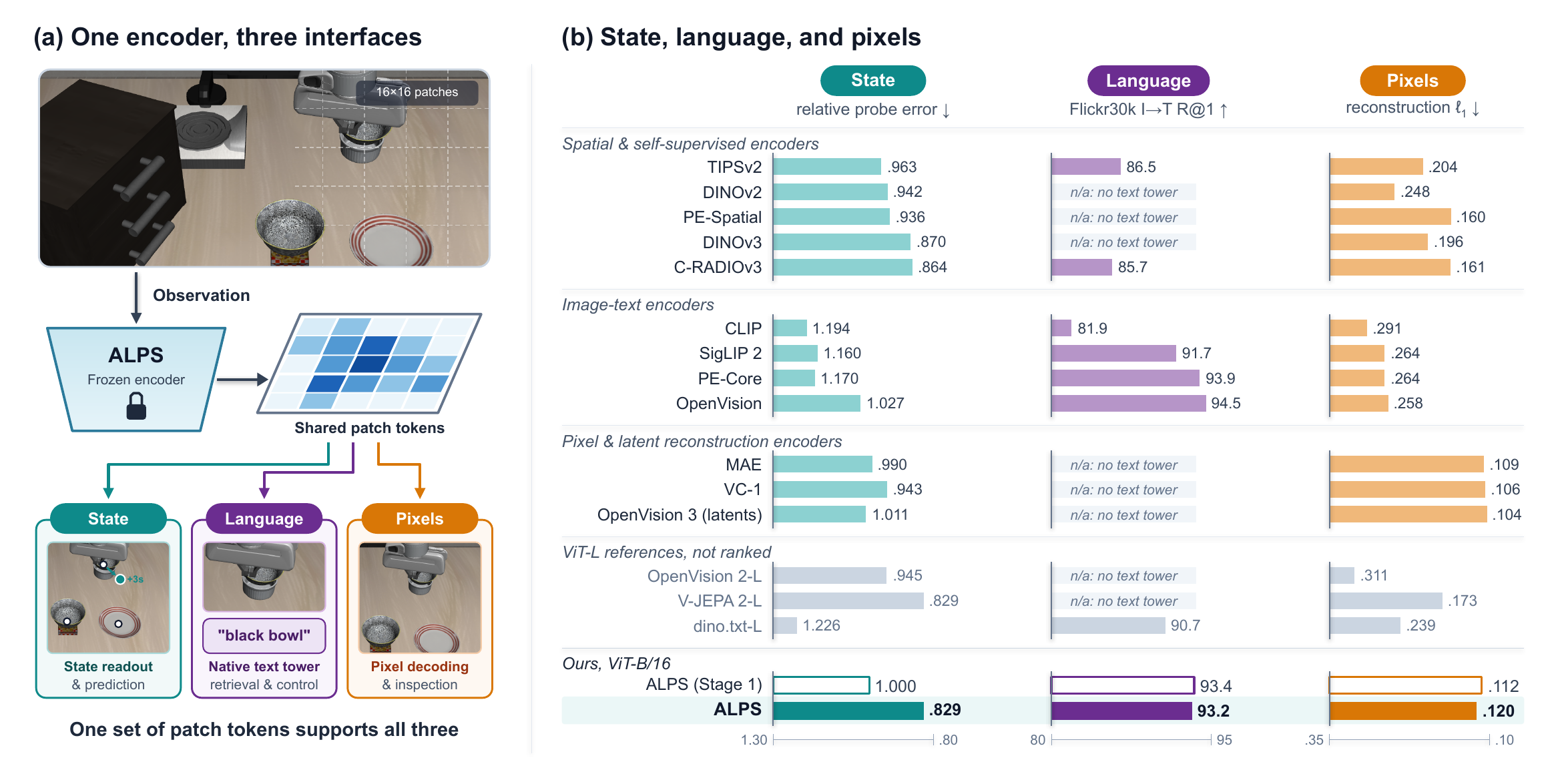}
\captionsetup{hypcap=false}
\captionof{figure}{\textbf{One frozen encoder for state, language, and pixels.}
(a) Conceptual illustration of the three interfaces of \ours{}.
(b) Every encoder of \cref{tab:main} on LIBERO state-probe error, Flickr30k image-to-text R@1, and the $\ell_1$ error of a decoder trained on its tokens, bars oriented so that longer is better; n/a: no text tower released (not zero).
\textbf{No compared encoder is better than \ours{} on all three.}}
\captionsetup{hypcap=true}
\label{fig:teaser}%
\end{strip}

%% file: sec/0_abstract.tex
\begin{abstract}
World models turn visual representations into predictions and actions.
We argue that their visual foundation should combine physical-state information, native language alignment, and pixel reconstruction, so that a single representation can support prediction, instruction, and inspection.
However, existing encoders typically specialize in only one or two of these capabilities; we therefore introduce \ours{}, a pretrained visual state encoder that unifies all three in a single set of patch tokens.
Using only still images and text, we pretrain a ViT-B/16 from scratch with language, reconstruction, and spatial correspondence objectives, and then refine it via cross-view self-distillation.
In a unified evaluation across fifteen public encoders, \ours{} achieves the lowest aggregate LIBERO state-probe error among ViT-B models (matching the $3.5\times$ larger V-JEPA~2-L) while maintaining strong language alignment and accurate pixel reconstruction.
Downstream, frozen \ours{} tokens reach the highest mean policy success on LIBERO (85.5\%) when paired with our native text tower, match or exceed leading encoders on world-model planning across PushT, Wall, and Granular, and allow predicted future tokens to be decoded directly into pixels for visual inspection.
We hope this work helps establish visual pretraining as a shared foundation for understanding, prediction, and action in world models.
\end{abstract}

%% file: sec/1_intro.tex
\section{Introduction}
\label{sec:intro}

World models predict how a scene changes in response to actions and use these predictions to plan.
Many of them operate in the latent space of a frozen pretrained visual encoder~\cite{zhou2024dino,baldassarre2025back,assran2025v}.
The encoder therefore decides what the world model can see: only when its tokens encode where objects are and what state they are in can a downstream world model learn how that state changes from scarce interaction data.
Two complementary interfaces make such an encoder far more versatile.
With tokens aligned to language, instructions and goals stated in words can be related to the scene without a second visual encoder.
With tokens that decode back into images, both observed scenes and imagined futures can be rendered and inspected.
This raises a central question: how can we pretrain a single visual encoder that provides physical-state information, language alignment, and pixel reconstruction at once?

Existing public encoders typically specialize in only one or two of these capabilities.
Self-supervised and agglomerative models such as DINOv2, DINOv3, and C-RADIOv3 provide transferable spatial features~\cite{oquab2023dinov2,simeoni2025dinov3,heinrich2025radiov2}, while predictive video encoders such as V-JEPA~2 learn spatiotemporal representations through latent prediction~\cite{assran2025v}.
Meanwhile, image-text encoders such as CLIP, SigLIP~2, and PE-Core align vision with language~\cite{radford2021learning,tschannen2025siglip,bolya2026perception}, and reconstruction-based encoders preserve fine visual detail in pixels or latents~\cite{he2022masked,zhang2026openvision}.
Combining state information, language alignment, and pixel reconstruction in a single representation is not straightforward.
A refinement that improves the state information of an image-text encoder, for example, can substantially degrade its language alignment, as we show for SigLIP~2 and CLIP (\cref{sec:stages}).

To resolve this tension, we pretrain our encoder in two stages (\cref{fig:method}).
The first stage establishes language alignment, pixel reconstruction, and spatial features in a shared representation.
We train a ViT-B/16 from scratch on image-text pairs with contrastive alignment, captioning, reconstruction, and dense correspondence objectives on the same patch tokens.
The second stage then strengthens state information through cross-view self-distillation, training the encoder to predict scene content across changes in crop, scale, and appearance.
From an augmented view, the student predicts a frozen first-stage teacher's patch tokens at corresponding locations in a clean view.
Both stages train on still images and text alone, without video or interaction supervision.
We call the final encoder \ours{} (\textbf{A} \textbf{L}anguage-aligned, \textbf{P}ixel-renderable \textbf{S}tate encoder) and the intermediate encoder after the first stage \ours{} (\stageone{}).

We compare \ours{} with fifteen public encoders, all frozen and evaluated under common protocols (\cref{fig:teaser}, \cref{tab:main}).
\ours{} achieves the lowest aggregate LIBERO state-probe error among ViT-B encoders, matching the $3.5\times$ larger V-JEPA~2-L, while retaining strong language alignment and accurate pixel reconstruction.
Cross-view refinement reduces state-probe error by 17\% while changing image-text retrieval by at most 0.5 points, and also improves world-model prediction.
In contrast, the same refinement lowers ImageNet zero-shot accuracy and image-to-text retrieval by 4--12 points for SigLIP~2 and CLIP (\cref{fig:refine}).
Frozen \ours{} tokens achieve the highest mean policy success in our LIBERO comparison when paired with our native text tower, and support leading planning performance on PushT, Wall, and Granular (\cref{sec:control}).
Predicted future tokens can also be decoded into images for inspection.

Our contributions are as follows:
\begin{itemize}
    \item We formulate visual state pretraining for world models and establish a unified evaluation protocol spanning physical-state information, native language alignment, pixel reconstruction, and downstream control, planning, and prediction.
    \item We introduce \ours{}, pretrained from scratch on still images and text with joint language, reconstruction, and spatial objectives followed by cross-view self-distillation.
    \item We demonstrate that one frozen representation can combine competitive performance across all three capabilities and serve as a shared visual foundation for robot policies and world models.
\end{itemize}

%% file: sec/2_related.tex
\section{Related Work}
\label{sec:related}

\begin{figure*}[t]
    \centering
    \includegraphics[width=\textwidth]{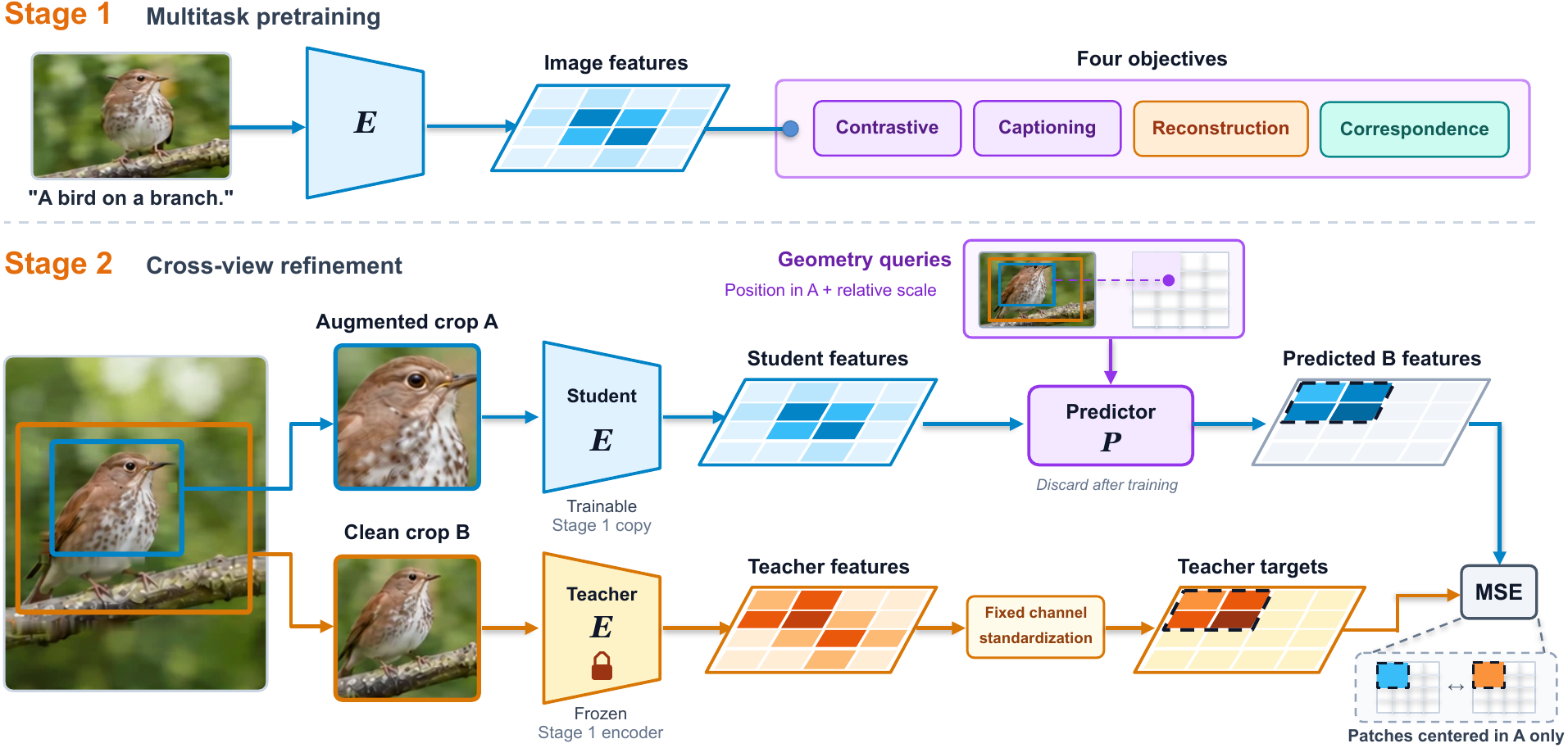}
    \caption{\textbf{Two-stage visual state pretraining.}
    \stageone{} trains an encoder $E$ from scratch with four objectives on shared patch tokens.
    \stagetwo{} initializes a student from $E$ and keeps $E$ frozen as the teacher.
    Given student tokens from an augmented crop $A$ and query tokens encoding position and relative scale, a predictor regresses standardized teacher tokens from a clean crop $B$.
    Only patches of $B$ whose centers lie inside $A$ contribute to the loss.
    After training, the predictor is discarded and the refined student serves as \ours{}.}
    \label{fig:method}
\end{figure*}

\paragraph{Image-text and unified visual pretraining.}
Contrastive image-text pretraining connects visual representations to language~\cite{radford2021learning,zhai2023sigmoid}.
Subsequent approaches incorporate captioning supervision~\cite{yu2022coca,li2025openvision,liu2026openvision}, self-distillation and masked prediction~\cite{tschannen2025siglip}, spatial objectives~\cite{maninis2025tips,cao2026tipsv2}, image reconstruction~\cite{tang2025tulip}, or autoregressive patch prediction~\cite{fini2025multimodal}.
Meanwhile, unified tokenizers and multimodal models couple understanding with generation~\cite{ma2026unitok,zhao2025qlip,wu2025vila}.
Other work instead adapts a frozen self-supervised backbone to language, as in dino.txt~\cite{jose2025dinov2}, or encodes the latents of a generative autoencoder, as in OpenVision~3~\cite{zhang2026openvision}.
Instead of adapting a frozen backbone or encoding autoencoder latents, our first stage jointly trains unnormalized patch tokens for language alignment, pixel reconstruction, and dense correspondence, providing a unified initialization for stage-two cross-view refinement.

\paragraph{Self-supervised and agglomerative representations.}
Self-distillation~\cite{caron2021emerging,zhou2021ibot,oquab2023dinov2,simeoni2025dinov3}, latent prediction~\cite{assran2023self,darcet2025cluster}, and multi-teacher distillation~\cite{ranzinger2024radio,heinrich2025radiov2} learn transferable patch representations.
To further strengthen dense spatial features, recent methods align region features with crop embeddings~\cite{wu2024clipself}, match patch-neighborhood orderings across views~\cite{pariza2025near}, enforce locality alignment~\cite{covert2025locality}, apply spatial refinement as in Perception Encoder~\cite{bolya2026perception}, or supervise predictions at multiple depths~\cite{mur2026v}.
Without relying on external teachers, our second stage refines the encoder from its own frozen first-stage weights through geometry-conditioned cross-view self-distillation, sharpening state information while preserving native language alignment and pixel reconstruction.

\paragraph{Visual representations for robot learning.}
Policies using pretrained visual representations can be competitive with those using ground-truth state inputs~\cite{parisi2022unsurprising}.
Robot representations have been learned from egocentric video~\cite{nair2022r3m}, masked autoencoding~\cite{radosavovic2023real,majumdar2023we}, language supervision~\cite{karamcheti2023language}, foundation-model distillation~\cite{shang2024theia}, and neural rendering~\cite{zhu2025spa}.
Comparative studies find no single dominant representation~\cite{majumdar2023we}, show that general vision datasets can match egocentric video for visuomotor pretraining~\cite{dasari2023unbiased}, and note that domain-specific pretraining does not guarantee robustness under visual shifts~\cite{burns2023makes}.
Pretrained on general-domain images and text, \ours{} provides frozen state tokens and a native text tower that transfer directly to language-conditioned manipulation.

\paragraph{Representations for world models.}
World models can learn a visual autoencoder before fitting latent dynamics~\cite{ha2018recurrent}, learn representations jointly with dynamics~\cite{hafner2025mastering,maes2026leworldmodel,kuang2026lpwm}, or build on pretrained features~\cite{baldassarre2025back}.
In particular, DINO-WM plans over frozen DINOv2 tokens~\cite{zhou2024dino}, V-JEPA~2 connects video pretraining to action-conditioned planning~\cite{assran2025v}, and frozen features also support language-conditioned planning~\cite{li2026grounded}, whereas RepWAM learns visual and action tokenizers directly from robot data~\cite{wang2026repwam}.
Generative world models instead simulate observations using learned tokenizers with autoregressive or diffusion dynamics~\cite{micheli2022transformers,bruce2024genie,agarwal2025cosmos}, or via pixel-space diffusion~\cite{alonso2024diffusion}.
Recent studies further show that off-the-shelf features do not automatically improve sample efficiency in model-based reinforcement learning~\cite{schneider2024surprising}, that pixel reconstruction fidelity alone is an unreliable proxy for recovering actions from latent transitions~\cite{yeom2026makes}, and that static image encoders can outperform video encoders depending on downstream protocols~\cite{terver2025drives}.
We focus on pretraining the visual encoder whose frozen tokens serve as the state representation for downstream world models.

%% file: sec/3_method.tex
\section{Method}
\label{sec:method}

Let $E$ map an image $x$ to patch tokens $z = E(x) \in \mathbb{R}^{N\times d}$.
Downstream state probes, policies, world models, and pixel decoders operate directly on the frozen tokens $z$.
For language alignment, a linear projection $W_{\text{img}}$ followed by average pooling and $\ell_2$ normalization maps $z$ to a global image embedding $v = \text{norm}\!\left(\frac{1}{N}\sum_{i=1}^{N} W_{\text{img}} z_i\right) \in \mathbb{R}^{D}$, which is paired with a text encoder $E_{\text{text}}$.

\subsection{\stageone{}: joint pretraining from scratch}
\label{sec:stage1}

\stageone{} builds a shared representation through complementary language, pixel, and spatial supervision (\cref{fig:method}, top).
Contrastive alignment and captioning connect the tokens to language, reconstruction preserves visual detail, and dense correspondence supervises local spatial relationships across views.

\paragraph{Architecture.}
The image encoder $E$ is a ViT-B/16~\cite{dosovitskiy2020image} with fixed sine-cosine position embeddings.
It outputs the raw patch tokens $z$ of the last transformer block without a final normalization layer.
We jointly train a text encoder and a caption decoder with the same width and depth as $E$ (\cref{sec:suppl_train}).
A lightweight convolutional decoder $D$ maps $z$ to pixels for the pretraining reconstruction objective, whereas all reported reconstructions use a separate evaluation decoder trained on frozen tokens (\cref{sec:setup}).

\paragraph{Objectives.}
Four objectives supervise the same patch tokens:
\begin{equation}
    \mathcal{L}_{1} = \mathcal{L}_{\text{itc}} + \mathcal{L}_{\text{cap}} + w_{\text{rec}}\,\mathcal{L}_{\text{rec}} + 0.25\,\mathcal{L}_{\text{dense}} .
\end{equation}

\emph{Contrastive alignment.}
Each image is paired with its web caption and one sentence from a longer synthetic caption.
We compute an InfoNCE loss~\cite{radford2021learning} between the mean-pooled, projected image tokens and each caption set, then average the two losses to obtain $\mathcal{L}_{\text{itc}}$.
Each loss uses its own softmax, so the two captions paired with the same image are never treated as negatives of each other.

\emph{Captioning.}
The caption decoder autoregressively predicts the full synthetic caption, conditioned on the image tokens and preceding text~\cite{yu2022coca}.

\emph{Reconstruction.}
The decoder $D$ reconstructs $x$ from both $z$ and a noisy copy of $z$.
The objective combines $\ell_1$, structural similarity, and image gradient terms (\cref{sec:suppl_train}).
We do not apply color augmentation to $x$, which directly serves as the reconstruction target.

\emph{Dense correspondence.}
This objective directly supervises patch correspondence across crops, scales, and flips.
We form a second view $g(x)$ via a random crop and flip of $x$, so the location of each patch in the original view is known.
A small projection head maps the tokens of both views to contrastive features.
For each patch of $g(x)$, the positive feature is sampled bilinearly from the projected token field of $x$ at the corresponding location.
Features at other locations in the same image and from other images serve as negatives in an InfoNCE loss $\mathcal{L}_{\text{dense}}$ (\cref{sec:suppl_train}), encouraging local discriminability.
Following prior work on dense correspondence and pixel-level consistency~\cite{wang2021dense,xie2021propagate}, this term adds explicit cross-view patch supervision to global image-text alignment and single-image reconstruction.

\paragraph{Data.}
We train on DataComp-1B~\cite{gadre2023datacomp} using its web captions and synthetic captions generated by the LLaVA-LLaMA-3 captioner from Recap-DataComp-1B~\cite{li2024if}, conditioned on the original web caption~\cite{liu2026openvision}.
Because the captioner uses a frozen CLIP ViT-L/14~\cite{li2024if}, knowledge from this image-text model reaches our encoder only through text; no external visual model supplies feature targets.
Training processes 12.8 billion samples at $128\times128$ resolution, followed by 1.0 billion samples at $224\times224$ resolution (\cref{sec:suppl_train}).
After \stageone{}, we retain the image encoder, text encoder, and image projection, while discarding the caption decoder, dense-correspondence projector, and $D$.

\begin{figure*}[t]
    \centering
    \includegraphics[width=\textwidth]{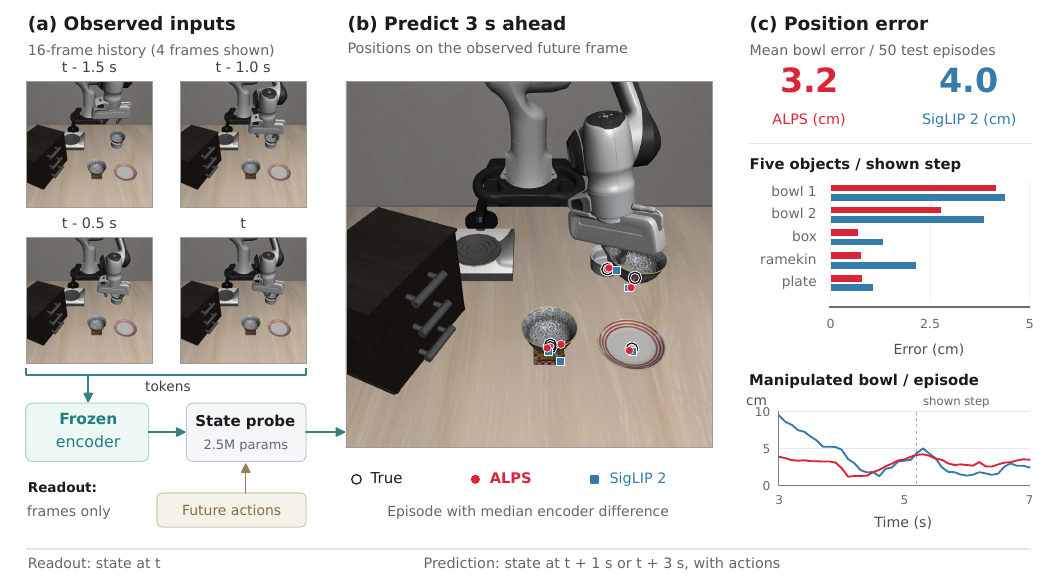}
    \caption{\textbf{The LIBERO state probe.}
    (a) A lightweight probe reads frozen tokens from the last 16 frames to estimate the current scene state from observations alone (readout) or predict the state 1 or 3\,s ahead given robot actions (prediction).
    (b) State predictions 3\,s ahead from \ours{} and SigLIP~2, overlaid on the frame observed 3\,s later, shown on the held-out LIBERO-Spatial episode with the median error difference between the two encoders.
    (c) Per-object position errors at this step, the manipulated bowl's error over the episode, and its mean error across 50 held-out episodes.}
    \label{fig:state}
    \vspace{-8pt}
\end{figure*}

\subsection{\stagetwo{}: cross-view self-distillation on single images}
\label{sec:stage2}

\stagetwo{} refines the shared representation through cross-view self-distillation (\cref{fig:method}, bottom).
From an augmented crop, the student predicts the tokens produced by a frozen \stageone{} teacher for a clean crop of the same image.
Known crop geometry specifies the target locations, and supervision focuses on the overlap between views.

\paragraph{Teacher and student.}
Both the frozen teacher and the trainable student are initialized from the \stageone{} encoder, and only the student encoder and predictor are updated.
Before training, we compute fixed per-channel statistics to standardize the teacher tokens throughout this stage.
For each image, we independently sample two views: the teacher receives a clean crop $B$ covering 40\% to 100\% of the image, while the student receives with equal probability a local crop $A$ (10\% to 40\%) or a global crop (40\% to 100\%) with photometric augmentation (\cref{sec:suppl_train}).

\paragraph{Geometry queries.}
Geometry queries specify where and at what scale the predictor should estimate teacher features.
A four-layer transformer predictor $P$ reads the student tokens $E(A)$ and produces one output for each query $q_i$ associated with patch $i$ of $B$.
In each layer, the queries first attend to one another and then to $E(A)$.
The crop boxes determine the position of each teacher patch center in the coordinate frame of $A$.
Each query encodes this position using a fixed sine-cosine embedding, measured in the patch units of $A$, and encodes the log side-length ratio of $B$ to $A$ through a learned embedding (\cref{tab:stage2_config}).
The queries thus supply the geometric correspondence, while the student features provide the visual content used to predict the teacher tokens.

\paragraph{Training objective.}
Let $V$ denote the set of teacher patches whose centers fall inside the student crop $A$.
We minimize the mean squared error between the predicted features and the standardized teacher tokens over $V$:
\begin{equation}
    \mathcal{L}_{2} = \frac{1}{|V|\,d} \sum_{i \in V} \big\| P\big(E(A), q_i\big) - \bar z^{\,\text{teacher}}_i(B) \big\|^2 .
\end{equation}
Patches outside $V$ are excluded from the loss, and we resample the student crop if $|V| < 16$.
Unlike masked prediction of unseen regions, this geometry-conditioned objective trains local patch tokens to remain consistent across changes in crop context, scale, and appearance while staying aligned with the frozen \stageone{} teacher.

\paragraph{Data and output.}
\stagetwo{} trains on 5 million single-image samples drawn from a curated collection of public image-editing datasets (\cref{sec:suppl_train}).
These sources provide high-resolution, object-centric scenes well-suited for local cross-view crops, which we sample as individual images without edit-pair or instruction supervision.
After training, we discard the predictor and retain the refined student encoder as \ours{}, paired directly with the unchanged \stageone{} text encoder and image projection.

%% file: sec/4_experiments.tex
\section{Evaluation protocol}
\label{sec:setup}

We evaluate the encoders across three core capabilities (physical-state information, language alignment, and pixel reconstruction; \cref{sec:intro}), as well as downstream control, planning, and world-model prediction.
All encoders remain frozen and follow a shared evaluation protocol on each benchmark.

\input{tab/main}

\paragraph{Encoders.}
We compare fifteen public encoders with our first-stage encoder and \ours{} (\cref{tab:main}).
Fourteen of the public models are image encoders, while V-JEPA~2 ViT-L~\cite{assran2025v} is a video encoder that takes the last 16 frames at $256\times256$ resolution and outputs tokens for the current frame.
Our primary comparison uses ViT-B image towers at $224\times224$ resolution (196 tokens for B/16 and 256 for B/14), taking each encoder's last-layer patch tokens with its default normalization.
Alongside the ViT-B models, OpenVision~2, V-JEPA~2, and dino.txt~\cite{jose2025dinov2} serve as ViT-L references; for dino.txt, we extract the patch tokens after its language-aligned vision head.

\paragraph{State probe.}
The state probe evaluates scene-state readout and action-conditioned prediction directly from frozen tokens (\cref{fig:state}).
Using demonstrations from LIBERO-Object and LIBERO-Spatial~\cite{liu2023libero}, a two-layer attentive probe (2.5M parameters) reads the tokens of the last 16 frames to estimate the movable objects' 3D positions, orientations, and binary predicates against simulator ground truth.
In the \emph{readout} setting, the probe estimates the current state from the frames alone; in the two \emph{prediction} settings, it additionally receives the robot actions over the next 1 or 3 seconds to predict the future state at that horizon.
We report position error relative to our first-stage encoder, averaged across the six combinations of the two suites and three settings (\cref{tab:libero} reports absolute errors in centimeters).
World models instead predict future tokens in latent space and are evaluated separately (\cref{sec:control}).

\paragraph{Language alignment.}
We evaluate language alignment using each model's native text tower, reporting ImageNet-1k~\cite{deng2009imagenet} zero-shot accuracy and Recall@1 for image-text retrieval on Flickr30k~\cite{young2014image} and COCO~\cite{lin2014microsoft}.
For encoders released without a paired text tower, these metrics are marked n/a, indicating the absence of a native text interface rather than a lack of semantic information in the features.
C-RADIOv3 uses the stronger of its two language adaptor heads, while \ours{} uses its \stageone{} language interface without modification.

\paragraph{Pixel reconstruction.}
To evaluate pixel reconstruction on equal footing, we train a separate lightweight convolutional adapter for each encoder that maps its frozen tokens to the latents of a frozen SDXL autoencoder~\cite{podell2024sdxl} for image decoding.
The adapter is trained on 19{,}000 ImageNet validation images and evaluated on 1{,}000 held-out images (one per class) using $\ell_1$ error and LPIPS~\cite{zhang2018unreasonable}.
Although much larger generative decoders can reconstruct plausible images from DINOv2 or SigLIP~2 tokens~\cite{zheng2026diffusion}, our protocol measures the visual detail directly recoverable by a lightweight probe, and \ours{} maintains its advantage when paired with a stronger decoder (\cref{sec:suppl_eval}).

\paragraph{Control.}
For visuomotor control, we train the LIBERO transformer policy~\cite{liu2023libero} on frozen visual tokens via behavior cloning with 50 demonstrations per task across four LIBERO suites.
Task instructions are embedded either by a frozen BERT model~\cite{devlin2019bert} or by each encoder's native text tower.
Each trained policy is evaluated three times over 50 rollouts per task (\cref{sec:suppl_policy}).

\paragraph{World-model prediction and planning.}
We follow the DINO-WM protocol~\cite{zhou2024dino} on PushT, Wall, PointMaze, Granular, and Rope, and LeWorldModel~\cite{maes2026leworldmodel} on Reacher.
On each task, we train an identical dynamics predictor over the frozen patch tokens of each encoder to support both cross-entropy planning and multi-step prediction evaluation (\cref{sec:suppl_eval}).

%% file: tab/main.tex
\begin{table*}[t]
\centering
\small
\setlength{\tabcolsep}{3.4pt}
\fitwidth{\textwidth}{%
\begin{tabular}{l cccc c cc cc cc}
\toprule
 & \multicolumn{4}{c}{State probe (LIBERO, relative error\,$\downarrow$)} & \multicolumn{5}{c}{Language alignment\,$\uparrow$} & \multicolumn{2}{c}{Reconstruction\,$\downarrow$} \\
\cmidrule(lr){2-5} \cmidrule(lr){6-10} \cmidrule(lr){11-12}
 & read- & & & & IN-1k & \multicolumn{2}{c}{Flickr30k R@1} & \multicolumn{2}{c}{COCO R@1} & & \\
Encoder & out & +1\,s & +3\,s & mean & 0-shot & I$\to$T & T$\to$I & I$\to$T & T$\to$I & $\ell_1$ & LPIPS \\
\midrule
\multicolumn{12}{l}{\emph{Spatial \& self-supervised encoders}} \\
TIPSv2-B/14~\cite{cao2026tipsv2}                  & .947 & .946 & .995 & .963 & 71.1 & 86.5 & 72.2 & 63.7 & 48.2 & .204 & .467 \\
DINOv2-B/14~\cite{oquab2023dinov2}                 & .916 & .901 & 1.010 & .942 & n/a & n/a & n/a & n/a & n/a & .248 & .535 \\
PE-Spatial-B/16~\cite{bolya2026perception}          & 1.000 & .892 & .917 & .936 & n/a & n/a & n/a & n/a & n/a & .160 & .386 \\
DINOv3-B/16~\cite{simeoni2025dinov3}                 & .860 & .911 & .840 & .870 & n/a & n/a & n/a & n/a & n/a & .196 & .424 \\
C-RADIOv3-B~\cite{heinrich2025radiov2}               & .876 & .838 & .877 & .864 & 66.5 & 85.7 & 74.3 & 56.5 & 46.0 & .161 & .359 \\
\midrule
\multicolumn{12}{l}{\emph{Image-text encoders}} \\
CLIP-B/16~\cite{radford2021learning}                & 1.091 & 1.110 & 1.380 & 1.194 & 68.4 & 81.9 & 62.0 & 52.5 & 33.1 & .291 & .579 \\
SigLIP~2-B/16~\cite{tschannen2025siglip}             & 1.149 & 1.202 & 1.130 & 1.160 & 77.6 & 91.7 & 78.1 & 67.9 & \textbf{52.4} & .264 & .541 \\
PE-Core-B/16~\cite{bolya2026perception}              & 1.118 & 1.099 & 1.292 & 1.170 & \textbf{78.4} & 93.9 & \textbf{80.9} & 71.1 & 50.3 & .264 & .559 \\
OpenVision-B/16~\cite{li2025openvision}              & 1.042 & 1.008 & 1.032 & 1.027 & 73.3 & \textbf{94.5} & 78.9 & \textbf{71.3} & 51.3 & .258 & .551 \\
\midrule
\multicolumn{12}{l}{\emph{Pixel \& latent reconstruction encoders}} \\
MAE-B/16~\cite{he2022masked}                         & .935 & 1.008 & 1.027 & .990 & n/a & n/a & n/a & n/a & n/a & .109 & .260 \\
VC-1-B~\cite{majumdar2023we}                         & 1.064 & .841 & .924 & .943 & n/a & n/a & n/a & n/a & n/a & .106 & .258 \\
OpenVision~3-B~\cite{zhang2026openvision}            & 1.029 & .934 & 1.071 & 1.011 & n/a & n/a & n/a & n/a & n/a & \textbf{.104} & \textbf{.243} \\
\midrule
\multicolumn{12}{l}{\emph{Larger encoders, ViT-L (304M to 330M parameters)}} \\
\textcolor{gray}{OpenVision~2-L/14~\cite{liu2026openvision}} & \textcolor{gray}{.969} & \textcolor{gray}{.881} & \textcolor{gray}{.987} & \textcolor{gray}{.945} & n/a & n/a & n/a & n/a & n/a & \textcolor{gray}{.311} & \textcolor{gray}{.602} \\
\textcolor{gray}{V-JEPA~2-L/16~\cite{assran2025v}}    & \textcolor{gray}{.741} & \textcolor{gray}{.841} & \textcolor{gray}{.904} & \textcolor{gray}{.829} & n/a & n/a & n/a & n/a & n/a & \textcolor{gray}{.173} & \textcolor{gray}{.413} \\
\textcolor{gray}{dino.txt-L/14~\cite{jose2025dinov2}} & \textcolor{gray}{1.090} & \textcolor{gray}{1.170} & \textcolor{gray}{1.418} & \textcolor{gray}{1.226} & \textcolor{gray}{81.6} & \textcolor{gray}{90.7} & \textcolor{gray}{75.8} & \textcolor{gray}{64.9} & \textcolor{gray}{46.7} & \textcolor{gray}{.239} & \textcolor{gray}{.552} \\
\midrule
\ours{} (\stageone{})                                & 1.000 & 1.000 & 1.000 & 1.000 & 74.5 & 93.4 & 79.4 & 70.6 & 52.0 & .112 & .365 \\
\rowcolor{oursrow}
\textbf{\ours{}}                                    & \textbf{.842} & \textbf{.834} & \textbf{.813} & \textbf{.829} & 73.4 & 93.2 & 78.9 & 70.2 & 51.9 & .120 & .372 \\
\bottomrule
\end{tabular}}
\caption{\textbf{Frozen encoders on state, language, and pixels: \ours{} has the lowest state-probe error among the ViT-B encoders.}
State probe: LIBERO position error relative to \ours{} (\stageone{}); language: each model's own text tower.
\textbf{Bold}: best ViT-B encoder in each column, \ours{} (\stageone{}) not ranked; \textcolor{gray}{gray}: ViT-L references; n/a: no released text tower (not zero).}
\label{tab:main}
\vspace{-8pt}
\end{table*}

%% file: sec/5_analysis.tex
\section{Results}
\label{sec:experiments}

\subsection{One encoder for state, language, and pixels}
\label{sec:main}

\begin{figure*}[t]
    \centering
    \includegraphics[width=\textwidth]{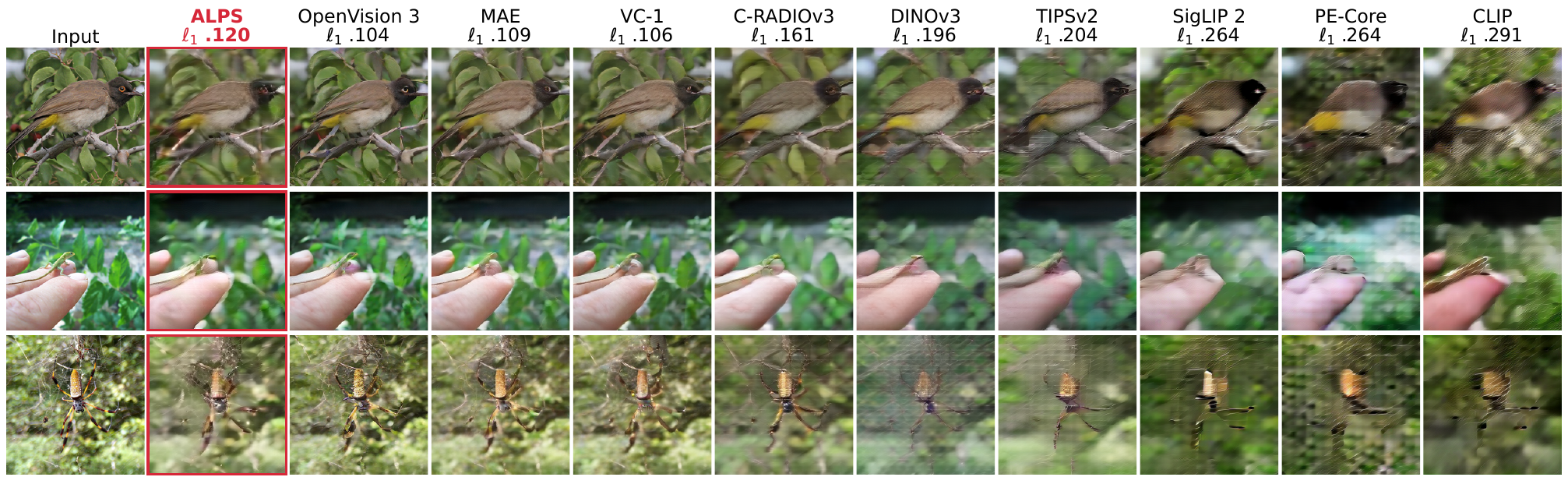}
    \caption{\textbf{Reconstruction from frozen tokens.}
    Following the input column, each column shows the evaluation decoder's output from one encoder's frozen tokens, labeled with its mean $\ell_1$ error across the 1{,}000 held-out ImageNet images.
    The rows show the first three images from a random sample (\cref{fig:recon_all}).
    \textbf{\ours{}, OpenVision~3, MAE, and VC-1 retain sharp edges and textures.
    C-RADIOv3 and DINOv3 preserve spatial layout with softer detail, while SigLIP~2, PE-Core, and CLIP mainly recover coarse structure.}}
    \label{fig:recon}
\end{figure*}

\Cref{tab:main} and \cref{fig:teaser}~(b) compare \ours{} with fifteen public encoders on state information, language alignment, and pixel reconstruction.

\paragraph{State.}
\ours{} achieves the lowest aggregate relative position error among the compared ViT-B encoders (.829), followed by C-RADIOv3 (.864) and DINOv3 (.870); together, these three models form a leading group ahead of PE-Spatial (.936).
\ours{} also achieves the lowest error in each of the three individual settings (readout, +1\,s, and +3\,s) as well as on frames rendered at $256\times256$ resolution (\cref{tab:libero_r256}).
While V-JEPA~2-L matches its aggregate error using $3.5\times$ as many parameters and video pretraining, the two models show complementary strengths: V-JEPA~2 reads out the current state more accurately (.741 vs.\ .842), whereas \ours{} is more accurate at 3-second prediction (.813 vs.\ .904).
Compared with SigLIP~2 and PE-Core (1.160 and 1.170), \ours{} reduces aggregate state-probe error by about 29\%, and probing their intermediate layers narrows this gap without closing it (\cref{sec:suppl_layers}).
On moving objects, \ours{} reaches a mean error of 3.22\,cm across the four prediction settings, matching the 3.04--3.25\,cm range of V-JEPA~2, DINOv3, and C-RADIOv3 (\cref{tab:libero_suppl}).

\paragraph{Language.}
\ours{} retains retrieval performance close to the strongest image-text encoders in our comparison.
It reaches 93.2/78.9 Recall@1 on Flickr30k and 70.2/51.9 on COCO (image-to-text/text-to-image), within 2 points of the best score on each of the four measures.
Its ImageNet zero-shot accuracy is 73.4, compared with 78.4 for PE-Core, the strongest ViT-B image-text encoder on this metric.
This larger 5.0-point gap on classification is expected when pretraining on descriptive synthetic captions, whose distribution differs from short ImageNet class prompts~\cite{liu2024clips}.

\begin{figure*}[t]
    \centering
    \includegraphics[width=\textwidth]{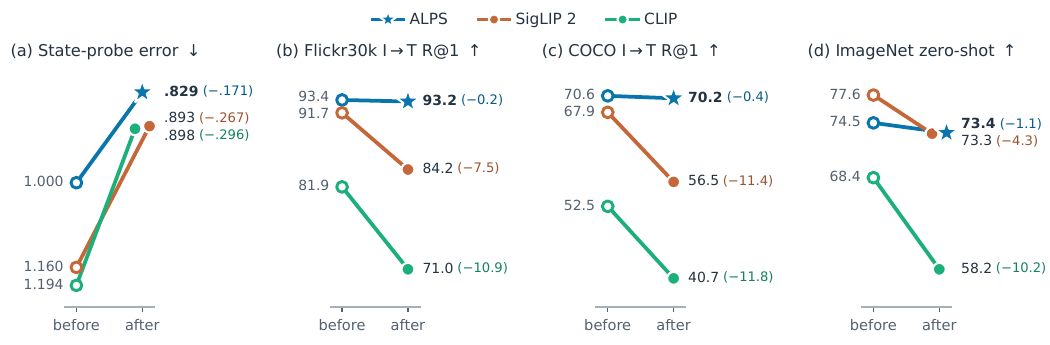}
    \caption{\textbf{The same \stagetwo{} refinement applied to \ours{} (\stageone{}), SigLIP~2, and CLIP}, each using its own frozen copy as the teacher.
    Lines connect each encoder before and after refinement, and the star marks \ours{}.
    Language is evaluated through each encoder's unchanged head.
    Higher values are better in every panel, with state-probe error relative to \ours{} (\stageone{}) inverted in (a).
    Endpoint labels show the value after \stagetwo{} and its change (\cref{tab:stage2_full}).
    \textbf{All three improve state information, while only \ours{} preserves language alignment.}}
    \label{fig:refine}
\end{figure*}

\paragraph{Pixels.}
\ours{} preserves fine visual detail that can be decoded directly from its tokens.
With the evaluation decoder, \ours{} achieves an $\ell_1$ error of .120 and an LPIPS of .372 (\cref{fig:recon}), yielding a lower $\ell_1$ error than every compared encoder not explicitly trained to reconstruct pixels or latents.
In particular, C-RADIOv3 has 34\% higher $\ell_1$ error (.161), DINOv3 has 63\% higher error (.196), and the image-text encoders have more than twice the error.
Encoders pretrained to reconstruct pixels (MAE at .109 and VC-1 at .106) or autoencoder latents (OpenVision~3 at .104) reconstruct more accurately, but incur 14\% to 22\% higher state-probe error.
LPIPS gives a slightly different ordering: these three reconstruction-focused encoders score .243--.260, and C-RADIOv3 also edges out \ours{} (.359, about 4\% lower).
Under a stronger decoder configuration, however, \ours{} maintains a lower $\ell_1$ error than C-RADIOv3 (.107 vs.\ .141, about 24\% lower) and overtakes it on LPIPS (.302 vs.\ .311; \cref{tab:recon_large}).

\paragraph{All three at once.}
\ours{} combines leading state-probe performance, strong language alignment, and accurate pixel reconstruction in one frozen representation.
Other leading state encoders have weaker retrieval or no paired text tower, while encoders with comparable retrieval or lower reconstruction error have higher state-probe error (\cref{fig:teaser}).
We use these interfaces for instruction-conditioned control and visual inspection of predicted world-model rollouts (\cref{sec:control}).

\subsection{Effect of cross-view refinement on state and language}
\label{sec:stages}

Cross-view refinement moves our encoder into the leading state group: before refinement, \ours{} (\stageone{}) has higher state-probe error than seven public ViT-B encoders in \cref{tab:main}, whereas after refinement it achieves the lowest.
\Cref{fig:refine} compares the same \stagetwo{} recipe across our first-stage encoder, SigLIP~2-B/16, and CLIP-B/16, each trained for 5 million samples with its own frozen copy as the teacher and evaluated through its unchanged language head (\cref{tab:stage2_full}).
For our encoder, refining on DataComp web images also reduces state-probe error, though less than our default curated collection (.886 vs.\ .829) under the same training budget, reflecting differences in both scene content and native resolution (\cref{sec:suppl_variants}).

\paragraph{All three encoders improve on state probing.}
\stagetwo{} reduces our encoder's aggregate state-probe error by 17\% (14\% to 19\% across the six individual settings) and raises DAVIS video object segmentation from .456 to .533 (\cref{sec:suppl_video}).
Similarly, SigLIP~2 improves from 1.160 to .893 on the state probe and from .503 to .552 on DAVIS, while CLIP improves from 1.194 to .898 and from .574 to .585, respectively.

\input{tab/policy}
\input{tab/planning}
\input{tab/predacc}

\paragraph{Only our encoder keeps its language alignment.}
For \ours{}, the improvement in state accuracy comes with minimal change to the other capabilities: image-text retrieval shifts by at most 0.5 points, ImageNet accuracy changes by about 1 point, and reconstruction remains close to its first-stage level ($\ell_1$ .120 vs.\ .112 and LPIPS .372 vs.\ .365).
In contrast, the public encoders lose substantially more language performance.
SigLIP~2 drops by 4.3 points on ImageNet accuracy and by 7.5 and 11.4 points on Flickr30k and COCO image-to-text retrieval, while CLIP drops by 10.2, 10.9, and 11.8 points, respectively.
These results show that cross-view refinement can strengthen state information while preserving the other interfaces, but this balance depends critically on the encoder being refined.

\paragraph{Lowering the learning rate does not remove the language gap.}
To test whether SigLIP~2 simply requires a smaller step size, we lower its encoder learning rate from $10^{-5}$ to $3 \times 10^{-6}$ while keeping all other settings unchanged.
Its state-probe error then falls by 17\% (from 1.160 to .959), matching the relative reduction of our encoder (from 1.000 to .829).
Yet it still loses more on every language metric: image-to-text Recall@1 falls by 2.3 points on Flickr30k and 2.9 on COCO, compared with 0.2 and 0.4 for \ours{} (\cref{sec:suppl_drift}).
Under the default refinement schedule, \ours{} also exhibits smaller drift in its patch tokens and image embeddings (\cref{sec:suppl_drift}).
A key structural difference lies in the language readout: \ours{} linearly projects mean-pooled patch tokens, whereas SigLIP~2 uses nonlinear attention pooling and CLIP reads a class token.
Because linear projection commutes with average pooling, zero-mean local token updates can cancel out in the pooled image representation, even without an explicit language constraint in \stagetwo{}.

\subsection{Control, planning, and prediction on frozen tokens}
\label{sec:control}

\input{tab/ablation}

We next test whether the learned representation supports downstream control, planning, and dynamics prediction.
Policies act directly on the encoded frames, while planners use a world model that predicts future tokens.

\paragraph{Policies.}
\ours{} achieves the highest mean policy success in our LIBERO evaluation (\cref{tab:policy}), which tests four suites using shared BERT instructions in the upper block and each encoder's native text tower in the lower block.
With BERT, \ours{} reaches 83.4\% mean success, outperforming spatial encoders such as C-RADIOv3 (78.2\%), DINOv2 (78.4\%), and DINOv3 (79.0\%) as well as SigLIP~2 (81.4\%).
Moreover, \stagetwo{} refinement raises LIBERO-Spatial success by 9.7 points, from 76.8\% to 86.5\% (per-suite breakdown in \cref{sec:suppl_policy}).

\paragraph{The text tower as the instruction interface.}
We replace BERT with projected embeddings from each encoder's native text tower, scaled to the mean norm of BERT embeddings.
Mean success rises from 83.4\% to 85.5\% for \ours{}, from 81.4\% to 84.8\% for SigLIP~2, from 78.2\% to 81.8\% for C-RADIOv3, and from 75.0\% to 79.2\% for CLIP.
Across three independent training runs per suite, \ours{} gains an average of 3.1 percentage points with its native text tower (\cref{sec:suppl_policy}).
These gains, especially on the multi-step instructions in LIBERO-Long, likely benefit from vision-language alignment during pretraining and, for \ours{}, joint training on detailed synthetic captions.

\paragraph{Planning.}
\ours{} leads or ties the best planning results on PushT, Wall, and Granular under the shared training protocols (\cref{tab:planning}).
It reaches 90\% success on PushT (tied with C-RADIOv3) and 92\% on Wall (tied with DINOv2), while all six encoders solve every goal on PointMaze.
On Granular, refinement substantially lowers the Chamfer distance from .324 (\stageone{}) to .176, ahead of DINOv3 and SigLIP~2.
On the remaining tasks, C-RADIOv3 achieves the lowest error on Rope (.558 vs.\ .750), and DINOv3 attains the highest success rate on Reacher (86.0\% vs.\ 78.7\%).

\paragraph{World-model prediction.}
Refinement also improves the accuracy of predicted states across four tasks (\cref{tab:predacc}).
We evaluate five-step token predictions using state probes trained on real-frame tokens, with true-token readout as a reference (\cref{tab:predacc_full}; protocol and rollout analysis in \cref{sec:suppl_eval}).
\stagetwo{} reduces the first-stage encoder's prediction error by 28\% on PushT, 17\% on Reacher, 13\% on Granular, and 11\% on Rope.
Compared with specialized spatial encoders, \ours{} outperforms DINOv3 on PushT and matches its accuracy on Granular and Rope, while C-RADIOv3 and DINOv3 achieve lower error on Reacher.
On Wall, true-token readout error is already close to prediction error, leaving little room to differentiate the encoders.

\begin{figure}[t]
    \centering
    \includegraphics[width=\linewidth]{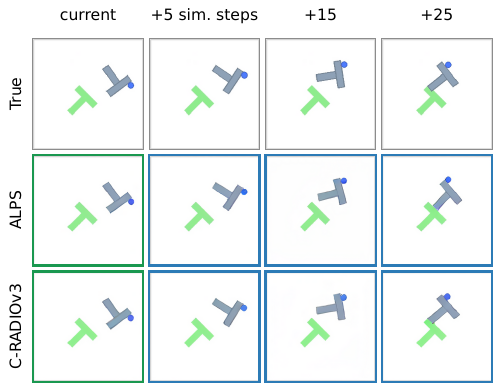}
    \caption{\textbf{Rendered world-model predictions on PushT} (window 834, largest true motion among 200 windows).
    The top row shows ground-truth frames; lower rows show each encoder's evaluation-decoder outputs from real current tokens (green) and open-loop predicted tokens given recorded actions (blue).
    Mean errors: \cref{tab:render}; full sequence: \cref{fig:render_w834}.}
    \label{fig:render_main}
\end{figure}

\paragraph{Rendering predicted tokens.}
The pixel interface makes the world model's imagined futures visible.
On PushT, an evaluation decoder trained only on real-frame tokens (\cref{sec:setup}) renders the world model's five-step open-loop predictions without participating in dynamics training or planning (\cref{fig:render_main,tab:render}).
The rendered rollouts clearly expose predicted object motion and visual errors, with \ours{} achieving lower five-step LPIPS (.020) than DINOv3 (.024), \stageone{} (.025), and SigLIP~2 (.031), behind only C-RADIOv3 (.018).

\subsection{Trade-offs between the first-stage objectives}
\label{sec:ablation}

\paragraph{State, language, and pixels.}
The first-stage objectives contribute differently to the three capabilities.
We train each variant from scratch at one tenth of the \stageone{} schedule, then apply the same \stagetwo{} refinement (\cref{tab:ablation}).
All variants retain contrastive alignment, so the ablation tests the three auxiliary objectives.
The reconstruction loss is essential for the pixel interface: the two variants without it incur $72\%$ and $78\%$ higher $\ell_1$ reconstruction error after \stageone{} (.194 and .200 vs.\ .113).
Removing reconstruction or captioning lowers state-probe error by $8.5\%$ and $4.3\%$, respectively, indicating a capacity trade-off among the auxiliary tasks.
Across all variants, \stagetwo{} reduces state-probe error by $14\%$ to $19\%$, largely preserves language performance, and changes $\ell_1$ error by less than $4\%$.
We therefore retain all objectives in the full recipe, trading a small amount of state-probe accuracy for accurate pixel decoding while still outperforming specialized ViT-B spatial encoders at scale (\cref{tab:main}).

\paragraph{Language retention across objectives.}
Refinement also preserves language alignment even without the three auxiliary objectives.
For the contrastive-only variant, ImageNet zero-shot accuracy changes from 69.8 to 67.9, while image-to-text Recall@1 changes from 91.0 to 89.3 on Flickr30k and from 67.1 to 64.1 on COCO.
Thus, the auxiliary objectives are not required to retain strong language alignment in this reduced-budget setting.

\paragraph{Dense correspondence.}
Dense correspondence directly supervises local patch matching across crops, scales, and flips (\cref{sec:stage1}).
Under the one-tenth budget of \cref{tab:ablation}, removing $\mathcal{L}_{\text{dense}}$ has minimal impact on global and single-view metrics (changing aggregate state-probe error by less than $1\%$, $\ell_1$ error by under $.001$, LPIPS negligibly, and retrieval by at most 1.1 points), as these benchmarks do not directly test cross-view patch correspondence.
When evaluated directly on cross-view patch matching (\cref{sec:suppl_ablation}), however, removing $\mathcal{L}_{\text{dense}}$ reduces matching accuracy by 6.0 points after \stageone{} (from 37.4\% to 31.4\%), and the full recipe retains a 2.6-point advantage even after \stagetwo{} refinement (40.8\% vs.\ 38.2\%).

%% file: tab/policy.tex
\begin{table}[b]
\centering
\small
\setlength{\tabcolsep}{3.5pt}
\fitwidth{\columnwidth}{%
\begin{tabular}{l cccc c}
\toprule
Encoder & Long & Goal & Spatial & Object & mean \\
\midrule
\multicolumn{6}{l}{\emph{Instruction embedded by BERT}} \\
CLIP-B/16                                      & 56.3 & 74.1 & 76.5 & 93.2 & 75.0 \\
SigLIP~2-B/16                                  & \textbf{72.3} & 78.8 & 84.1 & 90.3 & 81.4 \\
DINOv2-B/14                                    & 66.5 & 78.6 & 80.5 & 88.1 & 78.4 \\
DINOv3-B/16                                    & 68.4 & 72.7 & 80.7 & \textbf{94.4} & 79.0 \\
C-RADIOv3-B                                    & 66.0 & 79.9 & 74.9 & 92.3 & 78.2 \\
\addlinespace[3pt]
\ours{} (\stageone{})                          & 68.9 & 84.3 & 76.8 & 94.3 & 81.1 \\
\textbf{\ours{}}                               & 69.7 & \textbf{85.1} & \textbf{86.5} & 92.2 & \textbf{83.4} \\
\midrule
\multicolumn{6}{l}{\emph{Instruction embedded by the encoder's own text tower}} \\
CLIP-B/16                                      & 61.5 & 84.3 & 79.1 & 91.9 & 79.2 \\
SigLIP~2-B/16                                  & 73.3 & \textbf{90.2} & \textbf{84.1} & 91.7 & 84.8 \\
C-RADIOv3-B                                    & 73.4 & 89.5 & 82.9 & 81.4 & 81.8 \\
\addlinespace[3pt]
\textbf{\ours{}}                               & \textbf{78.5} & 87.4 & 82.6 & \textbf{93.7} & \textbf{85.5} \\
\bottomrule
\end{tabular}}
\caption{\textbf{Policy learning on LIBERO with frozen tokens: \ours{} achieves the highest mean success with both BERT and its native text tower.}
Success in \% (mean of three evaluations, 50 rollouts per task); instructions from BERT (top) or each encoder's own text tower (bottom). Bold: best in each block.}
\label{tab:policy}
\end{table}

%% file: tab/planning.tex
\begin{table*}[t]
\centering
\small
\setlength{\tabcolsep}{5pt}
\fitwidth{0.92\textwidth}{%
\begin{tabular}{l cc cc c c c c}
\toprule
 & \multicolumn{2}{c}{PushT (50)} & \multicolumn{2}{c}{Wall (50)} & PointMaze (50) & Reacher (50) & Granular (10) & Rope (50) \\
\cmidrule(lr){2-3} \cmidrule(lr){4-5} \cmidrule(lr){6-6} \cmidrule(lr){7-7} \cmidrule(lr){8-8} \cmidrule(lr){9-9}
Encoder & success\,$\uparrow$ & state dist.\,$\downarrow$ & success\,$\uparrow$ & state dist.\,$\downarrow$ & success\,$\uparrow$ & success\,$\uparrow$ & Chamfer\,$\downarrow$ & Chamfer\,$\downarrow$ \\
\midrule
DINOv2-B/14                                    & 88 & 25.0 & \textbf{92} & 3.62 & 100 & 76.7 & .211 & .701 \\
DINOv3-B/16                                    & 88 & 24.2 & 90 & 3.69 & 100 & \textbf{86.0} & .275 & .681 \\
C-RADIOv3-B                                    & \textbf{90} & \textbf{20.2} & 90 & 3.31 & 100 & 82.0 & .180 & \textbf{.558} \\
SigLIP~2-B/16                                  & 84 & 29.0 & 88 & 3.73 & 100 & 74.0 & .230 & .848 \\
\midrule
\ours{} (\stageone{})                          & 88 & 24.0 & 90 & 3.43 & 100 & 76.0 & .324 & .943 \\
\textbf{\ours{}}                               & \textbf{90} & 23.4 & \textbf{92} & \textbf{3.21} & 100 & 78.7 & \textbf{.176} & .750 \\
\bottomrule
\end{tabular}}
\caption{\textbf{Planning with world models trained on frozen tokens: \ours{} leads or ties the best results on PushT, Wall, and Granular.}
DINO-WM~\cite{zhou2024dino} on PushT, Wall, PointMaze, Granular, and Rope and LeWorldModel~\cite{maes2026leworldmodel} on Reacher, trained with one recipe (\cref{sec:suppl_eval}); in parentheses: goals evaluated; Reacher: mean of three planner seeds. Success in \%; state dist.: final-to-goal state distance; Chamfer: final-to-goal particle distance. Bold: best of a column; none if all tie.
}
\label{tab:planning}
\end{table*}

%% file: tab/predacc.tex
\begin{table*}[t]
\centering
\small
\setlength{\tabcolsep}{12pt}
{%
\begin{tabular}{l ccccc}
\toprule
 & PushT & Wall & Granular & Rope & Reacher \\
Encoder & block (units) & agent (units) & Chamfer & Chamfer & joints (deg) \\
\midrule
DINOv2-B/14                                    & 7.16 & \textbf{9.69} & .246 & .410 & 1.21 \\
DINOv3-B/16                                    & 8.70 & 9.78 & .222 & .407 & \textbf{1.03} \\
C-RADIOv3-B                                    & \textbf{5.64} & 9.75 & \textbf{.218} & \textbf{.372} & 1.06 \\
SigLIP~2-B/16                                  & 11.96 & 9.91 & .268 & .469 & 2.02 \\
\midrule
\ours{} (\stageone{})                          & 9.32 & 9.81 & .272 & .424 & 1.42 \\
\textbf{\ours{}}                               & 6.73 & \textbf{9.69} & .236 & .377 & 1.18 \\
\bottomrule
\end{tabular}}
\caption{\textbf{State read out from predicted tokens, 5 world-model steps ahead: \stagetwo{} reduces prediction error by 11--28\% across four tasks.}
A probe trained on real frames reads the world-model predictions of \cref{tab:planning} and \cref{tab:rope} on held-out trajectories; lower is better, positions in workspace units.
Bold: lowest of a column. Every step and two references: \cref{tab:predacc_full}; protocol: \cref{sec:suppl_eval}.}
\label{tab:predacc}
\end{table*}

%% file: tab/ablation.tex
\begin{table*}[t]
\centering
\small
\setlength{\tabcolsep}{3.4pt}
\fitwidth{\textwidth}{%
\begin{tabular}{l cccc c cc cc cc}
\toprule
 & \multicolumn{4}{c}{State probe (LIBERO, relative error\,$\downarrow$)} & \multicolumn{5}{c}{Language alignment\,$\uparrow$} & \multicolumn{2}{c}{Reconstruction\,$\downarrow$} \\
\cmidrule(lr){2-5} \cmidrule(lr){6-10} \cmidrule(lr){11-12}
 & read- & & & & IN-1k & \multicolumn{2}{c}{Flickr30k R@1} & \multicolumn{2}{c}{COCO R@1} & & \\
\stageone{} objectives & out & +1\,s & +3\,s & mean & 0-shot & I$\to$T & T$\to$I & I$\to$T & T$\to$I & $\ell_1$ & LPIPS \\
\midrule
\multicolumn{12}{l}{\emph{After \stageone{} at 10\% of its schedule}} \\
all four objectives        & 1.000 & 1.000 & 1.000 & 1.000 & 68.9 & 89.6 & 72.8 & 66.3 & 46.6 & .113 & .333 \\
no reconstruction          & .956 & .917 & .873 & .915 & 69.6 & 90.8 & 73.7 & 67.1 & 47.1 & .194 & .412 \\
no dense correspondence    & 1.010 & 1.019 & .994 & 1.008 & 69.1 & 88.5 & 73.4 & 66.0 & 46.5 & .113 & .345 \\
no captioning              & .978 & .940 & .953 & .957 & 67.9 & 89.6 & 72.5 & 66.2 & 45.9 & .112 & .336 \\
contrastive only           & .969 & 1.043 & 1.085 & 1.033 & 69.8 & 91.0 & 74.4 & 67.1 & 47.6 & .200 & .447 \\
\midrule
\multicolumn{12}{l}{\emph{After the \stagetwo{} refinement of each variant}} \\
all four objectives        & .838 & .822 & .932 & .864 & 67.0 & 88.4 & 71.4 & 65.3 & 45.8 & .116 & .343 \\
no reconstruction          & .751 & .772 & .721 & .748 & 67.1 & 89.7 & 72.2 & 64.5 & 46.5 & .196 & .400 \\
no dense correspondence    & .865 & .805 & .927 & .866 & 67.5 & 88.4 & 71.9 & 64.2 & 46.0 & .116 & .351 \\
no captioning              & .771 & .777 & .774 & .774 & 66.3 & 88.5 & 71.4 & 63.8 & 45.5 & .116 & .350 \\
contrastive only           & .822 & .840 & .948 & .870 & 67.9 & 89.3 & 72.9 & 64.1 & 46.8 & .205 & .433 \\
\bottomrule
\end{tabular}}
\caption{\textbf{\stageone{} objectives at one tenth of the budget.}
Each variant is pretrained from scratch with 10\% of the \stageone{} schedule and refined with \stagetwo{}; all keep the contrastive objective. State probe relative to the all-four variant after \stageone{}; other columns as in \cref{tab:main}.
\textbf{Reconstruction gives the pixel interface at a cost in state-probe error.}}
\label{tab:ablation}
\end{table*}

%% file: sec/6_conclusion.tex
\section{Conclusion}
\label{sec:conclusion}

\ours{} shows that a single frozen visual representation, pretrained on still images and text, can combine competitive state information, native language alignment, and pixel reconstruction.
Cross-view refinement strengthens state representations while preserving the language and pixel interfaces.
The same tokens support robot control, world-model prediction and planning, and visual inspection of predicted scenes.
These results suggest visual state pretraining as a shared foundation for robot policies and world models.

\paragraph{Limitations and future work.}
We focus on a ViT-B/16 backbone and simulation benchmarks for controlled comparison across fifteen encoders, leaving larger model scales and real-robot deployment to future work.
On the training side, our recipe relies on a two-stage curriculum to balance reconstruction, language, and spatial state (\cref{sec:ablation}); folding cross-view refinement into a single training stage remains an open question.
Moreover, \ours{} is pretrained without video supervision and is less sensitive to fine-grained temporal actions despite improving cross-frame correspondence on DAVIS (\cref{sec:suppl_video}), which motivates extending \stagetwo{} to video and interaction data.
Finally, our downstream evaluations test language-conditioned control and world-model planning in separate settings; integrating state, language, and pixel decoding into a single language-guided, pixel-verifiable planner is a natural next step.

%% file: sec/X_suppl.tex
\setcounter{page}{1}
\renewcommand{\thepage}{S\arabic{page}}
\maketitlesupplementary

\setcounter{section}{0}\setcounter{table}{0}\setcounter{figure}{0}
\renewcommand{\thesection}{S\arabic{section}}\renewcommand{\thetable}{S\arabic{table}}\renewcommand{\thefigure}{S\arabic{figure}}
\renewcommand{\theHsection}{S\arabic{section}}\renewcommand{\theHtable}{S\arabic{table}}\renewcommand{\theHfigure}{S\arabic{figure}}   %

\renewcommand{\dblfloatpagefraction}{0.7}\renewcommand{\floatpagefraction}{0.7}

\section{Training details}
\label{sec:suppl_train}

\paragraph{\stageone{}.}
\Cref{tab:stage1_hparams} lists the architecture and optimization hyperparameters of \stageone{}.
Both the text encoder and the caption decoder have 12 layers and width 768; a class token participates in attention but is not used in any loss.
Training runs at $128\times128$ resolution for 12.8 billion samples (batch size 32{,}768, peak learning rate $1.0 \times 10^{-3}$, $w_{\text{rec}} = 2$) and then at $224\times224$ resolution for 1.0 billion samples (batch size 16{,}384, peak learning rate $2.6 \times 10^{-5}$, $w_{\text{rec}} = 1$).
The contrastive temperature is learned (initialized at 0.07), with the logit scale clamped between 1 and 100.
At each step, the second positive caption for each image is a single randomly sampled sentence from its synthetic caption, whereas the caption decoder predicts the full synthetic caption.

\emph{Reconstruction.} With $\tilde z = z + 0.05\,\mathrm{rms}(z)\,\epsilon$ and $\epsilon \sim \mathcal{N}(0, I)$, where the per-token root mean square $\mathrm{rms}(z)$ is computed without backpropagating gradients, the reconstruction loss is $\mathcal{L}_{\text{rec}} = \rho\big(D(z), x\big) + 0.25\,\rho\big(D(\tilde z), x\big)$, where $\rho$ combines $\ell_1$, structural similarity, and image gradient terms with weights 1, 0.2, and 0.1.

\emph{Dense correspondence.} The projection head maps the tokens of both views to 256 dimensions, and $\mathcal{L}_{\text{dense}}$ is an InfoNCE loss with temperature 0.1. It is computed as a weighted sum of two separate softmax terms, one over negatives from other locations in the same image (weight 0.75) and one over negatives from other images (weight 0.25), which keeps their balance fixed across resolutions.
At 128 pixels, the negatives consist of all 63 other locations in the same image and 64 locations from other images; at 224 pixels, we sample 64 negatives of each type while excluding the $3 \times 3$ neighborhood around the positive location.
After training, we retain the image encoder, text encoder, and contrastive image projection, and discard the caption decoder, dense-correspondence projector, and pretraining decoder.

\input{tab/stage1_hparams}

\paragraph{\stagetwo{}.}
\looseness=-1 %
\Cref{tab:stage2_config} lists the crop views, predictor architecture, and target settings of \stagetwo{} for \ours{}, matching our training configuration and code.
The query for patch $i$ of the teacher crop $B$ is defined as
\begin{gather*}
q_i = m + \mathrm{PE}(p_i) + W_r \big[\ln r,\ (\ln r)^2\big], \\
p_i = 14 \Big(\frac{y_i - y_A}{s_A},\ \frac{x_i - x_A}{s_A}\Big), \qquad r = \frac{s_B}{s_A},
\end{gather*}
where $m$ is a learned embedding, $\mathrm{PE}$ is the fixed sine-cosine positional encoding, $W_r$ is a learned linear projection, $(y_A, x_A)$ is the top-left corner of the student crop $A$, $s_A$ and $s_B$ are the side lengths of $A$ and $B$, and $(y_i, x_i)$ is the center of patch $i$ in $B$, all measured in the coordinate frame of the image's central square; $p_i$ is thus expressed in patch units of $A$.
\stagetwo{} trains for 4{,}883 steps at batch size 1{,}024 (5M sampled images in total), with a learning rate of $10^{-4}$ for the predictor and $10^{-5}$ for the encoder.
We use AdamW ($\beta_1 = 0.9$, $\beta_2 = 0.99$) with weight decay 0.05, gradient clipping at 1.0, 244 warm-up steps, and a cosine learning-rate schedule.
Photometric augmentation on the student crop applies color jitter (brightness 0.4, contrast 0.4, saturation 0.2, no hue shift) with probability 0.8, grayscale conversion with probability 0.1, and Gaussian blur ($\sigma \in [0.1, 2.0]$ pixels) with probability 0.5.

\input{tab/stage2_config_suppl}

\paragraph{\stagetwo{} training data.}
\stagetwo{} trains on 5M single-image samples drawn from a curated collection of public image-editing datasets (\cref{tab:stage2_sources}), which provide high-resolution, object-centric scenes.
Dataset metadata and edit pairings are used only for deduplication, filtering, and sampling weights.
Specifically, we remove exact duplicates; global or camera-level edits (style, tone, weather, lighting, background, or viewpoint) and edits of unknown type; image pairs that are identical, misaligned, unchanged, or re-rendered under a patch-wise pixel check at $224\times224$ resolution; and frames from real videos.
We cap each instruction group at 200 instances and each source image at 4 instances, applying no further filtering on resolution, aesthetics, or content.
During training, attribute-level records are subsampled to 10\% of the samples, and physical-state records (relations, states, and poses) are sampled with probability 0.05.
Each training sample selects a record according to these weights and takes either its source or edited image with probability 0.5; both crops $A$ and $B$ are extracted from this single image, and neither the edit pairing nor any text instruction is used in the training objective.
Every refinement experiment in this paper uses this exact budget of 5M sampled images (4{,}883 steps at batch size 1{,}024).
The collection contains no real video: at most 1.8\% of the sampled images are individual frames from a generative video dataset (PICA-100K), and at most 2.5\% come from AURORA, which mixes edited photographs with simulator renderings.

\input{tab/stage2_sources}
\input{tab/libero}
\input{tab/libero_suppl}

\section{Evaluation details}
\label{sec:suppl_eval}

\paragraph{State probe.}
\Cref{fig:state} in the main text illustrates the state-probe setup; \cref{tab:libero} reports per-setting position errors in centimeters, and \cref{tab:libero_suppl} reports the remaining outputs of the probe, which jointly predicts positions, orientations, and binary predicates.

\emph{Data.} We use agent-view camera recordings from the LIBERO-Object and LIBERO-Spatial demonstrations (10 tasks with 50 demonstrations each), flipped upright and subsampled from 20 to 10\,Hz; frames are stored at $128\times128$ resolution and resized to $224\times224$ via bicubic interpolation.
For each task, demonstrations are split by episode into 35 for training, 5 for validation, and 5 for testing, while the remaining 5 are held out as a reserve test set unused in this work.
Every frame whose prediction target falls within the same episode serves as a valid sample, yielding roughly 11k--26k training samples and 1.5k--3.7k test samples per setting.

\emph{Targets.} Each movable object in a suite is assigned a fixed slot in alphabetical order: 11 slots in LIBERO-Object (7 present per task) and 5 in LIBERO-Spatial, with absent objects masked out.
From the simulator ground truth, we extract each object's 3D world-frame position, its orientation (predicted as a 6D rotation representation and evaluated up to object symmetries), and the suite's 10 binary \emph{on} and \emph{in} predicates.
The prediction settings target the scene state 10 or 30 frames ahead while supplying the probe with the 20 or 60 intervening robot actions.

\emph{Probe.} Patch tokens from the last 16 frames are standardized per channel using statistics computed from every fifth training episode, then projected to width 256 with learned spatial and temporal embeddings; robot actions are embedded via a lightweight MLP.
Eight learned queries pass through two transformer decoder layers (4 heads, feed-forward width 1{,}024) that cross-attend to all visual and action tokens, followed by linear heads that predict positions, rotations, and predicates.
The training objective sums the $\ell_1$ position loss, $0.1\times$ the rotation loss, and the binary cross-entropy predicate loss.
We train the probe using AdamW (learning rate $3\times10^{-4}$, weight decay 0.01, cosine schedule without warm-up, gradient clipping at 1.0) for 6{,}000 steps with batch size 128 and bfloat16 mixed precision, evaluating validation loss every 1{,}000 steps and selecting the checkpoint with the lowest validation loss.
We report mean Euclidean position error across all test samples and present objects.
Two simple baselines contextualize these errors: predicting each object's mean training position yields 9.6 to 16.1\,cm error across the six settings, whereas copying the true current position yields 1.5 to 4.8\,cm overall at +1 and +3\,s (since most objects remain static) but 12.8 to 31.6\,cm on moving objects, where all tested encoders achieve under 6\,cm (\cref{tab:libero_suppl}).

\emph{Encoders.} Each encoder uses its default input normalization; replacing DINOv2's native normalization with the DINO-WM convention (mean and standard deviation 0.5) lowers its relative error from .942 to .909.
V-JEPA~2 takes 16-frame clips at $256\times256$ resolution, whereas TIPSv2 and PE-Spatial are evaluated at $224\times224$ rather than their native 448 and 512 resolutions.
For MAE (the pretrained ViT-B/16 without fine-tuning), VC-1 (the base release), and PE-Spatial, we extract patch tokens after the final normalization layer.
For dino.txt, both the state probe and evaluation decoder read the patch tokens after its vision head (256 tokens of width 1{,}024), while language tasks use its default image representation (the class token concatenated with the mean-pooled patch token after the vision head), tokenizer, and preprocessing.
Using the tokens after its vision head, dino.txt yields the highest state-probe error in our comparison (1.226); extracting tokens directly from its frozen DINOv2 ViT-L backbone (which lacks language alignment) lowers the error to .924, matching DINOv2 ViT-B (.942).

\paragraph{Frames rendered at $256\times256$ resolution.}
\label{sec:suppl_r256}

\input{tab/libero_r256}

Standard LIBERO demonstrations store frames at $128\times128$ resolution, which are upsampled to each encoder's input size.
To verify that our findings are not an artifact of low-resolution storage, we re-render the camera frames of all episodes directly at $256\times256$ from the recorded simulator states and repeat the state-probe evaluation across all encoders in \cref{tab:main} (\cref{tab:libero_r256}), resizing from 256 pixels to each encoder's standard input resolution.
At $128\times128$, re-rendering matches the stored dataset frames within a mean absolute difference of 3.2 out of 255 gray levels.
Higher native resolution reduces position error across all models, and \ours{} again achieves the lowest mean relative error among ViT-B encoders (.845), followed by DINOv3 (.862) and C-RADIOv3 (.895), while V-JEPA~2 ViT-L reaches .856.
CLIP, SigLIP~2, PE-Core, OpenVision~3, dino.txt, and MAE benefit the most from higher resolution (e.g., SigLIP~2 improves from 1.160 to 1.066 and PE-Core from 1.170 to 1.078, relative to our first-stage encoder on the same frames), yet still trail the leading spatial encoders.

\paragraph{Intermediate layers.}
\label{sec:suppl_layers}

\input{tab/layers}

Because image-text encoders often retain stronger spatial structure in intermediate layers than in their final output~\cite{bolya2026perception}, evaluating only the last layer in \cref{tab:main} could understate their spatial capability.
We therefore probed blocks 6, 8, and 10 (out of 12, taking unnormalized block outputs) for SigLIP~2 and PE-Core using the same protocol, selecting for each encoder the block with the lowest mean validation loss across the six settings according to a pre-specified rule.
Both encoders select block 6 (\cref{tab:layers}), lowering aggregate relative error to 1.147 for SigLIP~2 and 1.009 for PE-Core, compared with 1.160 and 1.170 at the final block; even against their best validation layer, the last layer of \ours{} reduces error by 28\% and 18\%, respectively.
No probed intermediate layer of either model matches \ours{}; the lowest test error across all probed blocks (.985 at block 8 of PE-Core) still exceeds that of \ours{} (.829).

\paragraph{Policy learning.}
\label{sec:suppl_policy}

\input{tab/policy_detail}
\input{tab/pertask_suppl}

\begin{figure*}[tp]
    \centering
    \includegraphics[width=\textwidth]{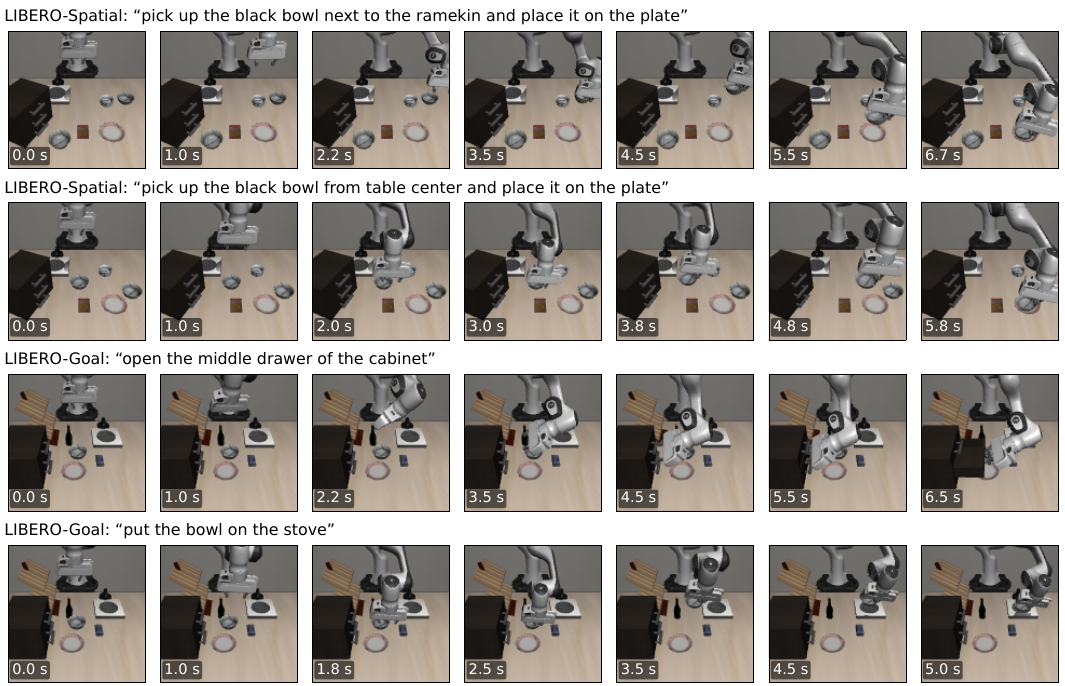}
    \caption{\textbf{Closed-loop rollouts of policies trained on frozen \ours{} tokens} across two tasks each from LIBERO-Spatial and LIBERO-Goal, shown with their language instructions.
We display the first two tasks (in canonical task order) whose recorded rollout succeeded; overall, 15 of the 20 recorded rollouts across the two suites (one per task) succeeded.}
    \label{fig:policy_all}
\end{figure*}

Simulator camera images are stored upside down and flipped upright prior to encoding.
Each success rate in \cref{tab:policy} is the mean across three evaluation seeds of a trained policy (50 rollouts per task per evaluation); \cref{tab:policy_detail} reports 95\% bootstrap confidence intervals over tasks and initial states, and \cref{tab:pertask} provides per-task success rates for the BERT-conditioned policies.
These intervals capture evaluation variance for fixed policy checkpoints across 4{,}000 bootstrap resamples of tasks and initial states; variance across independent policy trainings is analyzed below.
With BERT instructions, \ours{} achieves 85.1\% on LIBERO-Goal and 86.5\% on LIBERO-Spatial, outperforming all five public baselines on both suites (72.7\%--79.9\% on Goal and 74.9\%--84.1\% on Spatial).
On LIBERO-Object and LIBERO-Long, \ours{} reaches 92.2\% and 69.7\%, performing on par with SigLIP~2, DINOv2, DINOv3, and C-RADIOv3 (88.1\%--94.4\% on Object and 66.0\%--72.3\% on Long), whereas CLIP reaches 93.2\% on Object but drops to 56.3\% on Long.
Notably, our first-stage encoder already reaches 84.3\% on Goal.
Under BERT conditioning, the worst-case single-task success rate of \ours{} is 71\% on Goal and 73\% on Spatial, compared with at most 63\% and 62\% for the public encoders, and 79\% on Object (where DINOv3 reaches 89\%).
\Cref{fig:policy_all} visualizes policy rollouts on two tasks per suite.

The upper block of \cref{tab:policy} embeds task instructions using a frozen BERT model.
To evaluate a self-contained visual-language system, we replaced BERT with the projected text embeddings from our native text tower (rescaled by a single global scalar so that their mean norm across the 40 instructions matches that of BERT) and retrained the policies with all other settings unchanged.
To account for training stochasticity, we ran three independent training seeds per suite for each text embedding and evaluated every resulting policy three times as in \cref{tab:policy}.
Averaged across the four suites and three training seeds, our native text tower achieves 85.2\% success versus 82.1\% with BERT, a paired improvement of 3.1 percentage points; hierarchical bootstrapping over training runs, tasks, and rollouts yields a 95\% confidence interval of $[-0.9, +7.1]$ points, showing that our native text tower serves as an effective instruction interface without an external BERT model.
The lower block of \cref{tab:policy} (and the indented rows of \cref{tab:policy_detail}) applies the same native-text replacement to each baseline that provides a text tower, keeping all training hyperparameters matched to its BERT counterpart; for \ours{}, these entries correspond to the first seed of the three runs above.
SigLIP~2 and CLIP use their own text towers, while C-RADIOv3 uses the text tower of the DFN CLIP ViT-H/14 teacher to which its language adaptor is aligned (matching its language interface in \cref{tab:main}); in each case, the projected text embeddings are globally scaled to match the mean BERT norm across the 40 instructions.
As with \ours{}, we compare the two instruction interfaces for each encoder via their mean success across the four suites.

\input{tab/policy_hparams}

\Cref{tab:policy_hparams} summarizes the policy hyperparameters; within each suite, all encoders and instruction interfaces share identical training and evaluation settings, including random seeds.
Complete reproduction configurations specifying seeds, training budgets, checkpoint selection, and evaluation parameters will be released with our code.
Relative to the default LIBERO implementation, we replace the end-to-end trained ResNet with a frozen ViT encoder and an attention pooling head, train LIBERO-Long for 25 epochs instead of 50, and evaluate the final checkpoint three times with 50 rollouts per task rather than selecting the best intermediate checkpoint.

\paragraph{Planning.}
\begin{figure}[t]
    \centering
    \includegraphics[width=\linewidth]{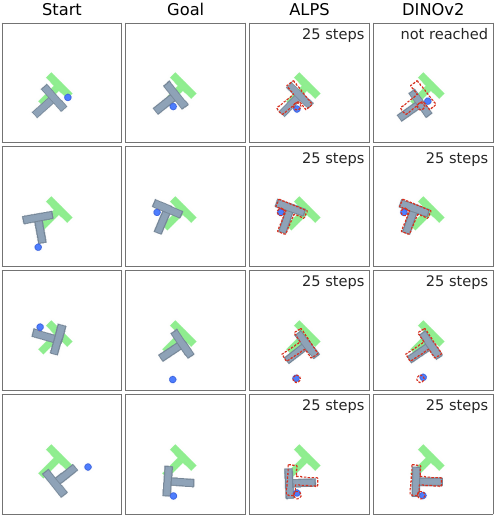}
    \caption{\textbf{Planning on PushT with a world model over frozen tokens}: start, goal, and final states for each encoder on the first four goals of \cref{tab:planning}.
Labels indicate executed simulator steps; dashed outlines mark the target block and agent poses.
This visualization comes from a separate run of the same CEM planner; because action sampling is stochastic, DINOv2 misses goal 1 here despite succeeding in the main evaluation of \cref{tab:planning}.}
    \label{fig:planning}
\end{figure}

\begin{figure*}[t]
    \centering
    \includegraphics[width=0.92\textwidth]{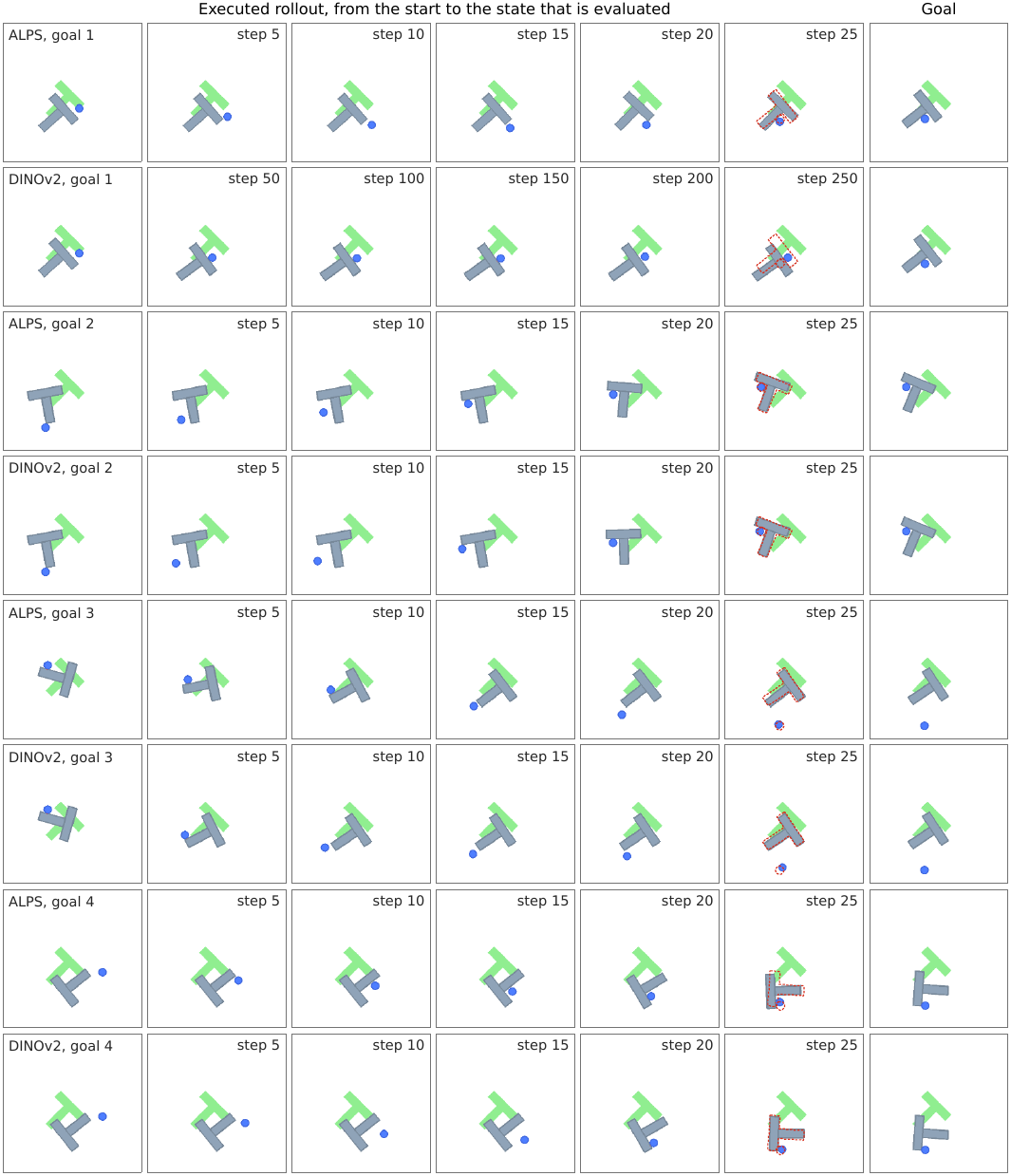}
    \caption{\textbf{Executed PushT rollouts for the four goals in \cref{fig:planning}}, comparing \ours{} and DINOv2: six frames from the initial state to the final evaluated state, followed by the goal image (dashed outlines mark the target block and agent poses).
DINOv2 misses goal 1 in this rollout (\cref{fig:planning}).}
    \label{fig:planning_strips}
\end{figure*}

\emph{World models.}
For PushT, Wall, PointMaze, Rope, and Granular, we adopt the codebase, datasets, and planner of DINO-WM~\cite{zhou2024dino}.
Each encoder is kept frozen, and its patch tokens are standardized per channel using training-set statistics.
The dynamics predictor is a 6-layer transformer that maps input tokens to predicted future tokens via learned linear projections, without training an image decoder.
All world models share a unified training recipe: batch size 256, learning rate $1.41 \times 10^{-3}$ (scaled by $\sqrt{256/32}$ from DINO-WM's base rate of $5 \times 10^{-4}$ at batch size 32), 200 warm-up steps, cosine decay to 1\% of the peak rate, and a constant 1\% floor over the final 5\% of training.
Total training updates scale with dataset size: 31{,}000 steps on PushT (roughly four passes over its training windows), 15{,}500 on Wall and PointMaze, and 7{,}750 on Rope and Granular.
Conditioning history consists of 3 frames with frame skip 5 on PushT and PointMaze, 1 frame with frame skip 5 on Wall, and 1 frame without frame skip on Rope and Granular.
Following DINO-WM, the predictor operates on a $14 \times 14$ token grid, so input resolution scales with patch size: $224\times224$ for patch size 16 and $196\times196$ for patch size 14 (DINOv2).
Every encoder thus produces exactly 196 patch tokens per frame across all five tasks.

\emph{Planning and evaluation.}
All five tasks use the Cross-Entropy Method (CEM) with 300 candidate sequences and 30 elites over a horizon of 5 world-model steps (30 optimization iterations on PushT and 10 on the remaining tasks), replanning after each execution round.
On PushT, one world-model step corresponds to 5 simulator steps; after executing the 25 planned simulator steps in a round, the planner replans from the newly reached state for up to 10 rounds.
A PushT goal is counted as reached when the combined agent and block position error falls below 20 units (in a $512\times512$ workspace) and the block orientation error is under 20 degrees.
PushT and Wall are evaluated across 50 goals, with Wall success defined as reaching within 4.5 units of the goal position.
PointMaze is evaluated on 50 random start-goal pairs, with success defined as reaching within 0.5 units of the target.
Following DINO-WM, Granular is evaluated on 10 instances and Rope on 50 instances using the Chamfer distance between final and goal particle states (where the first 10 Rope instances match DINO-WM's default evaluation seed).
Goal states are generated by rotating the initial rope by up to 90 degrees with translation, or by scaling the initial particle pile by a factor of 0.6--0.9 with translation.
Neither task defines an early-stopping success threshold, so planning always executes all 10 rounds.
All encoders are evaluated on identical goals using the same planner seed.
\Cref{fig:planning} illustrates planning results on PushT, and \cref{fig:planning_strips} shows the corresponding rollout trajectories.

\input{tab/rope_suppl}

\emph{Rope.}
\Cref{tab:rope} supplements the mean Chamfer distance in \cref{tab:planning} with the median across all 50 instances.
As a pipeline verification, DINOv2-S/14 trained with DINO-WM's original predictor and schedule achieves .567 on the 10 default DINO-WM instances (compared with .41 reported in DINO-WM) and .547 across all 50 instances.

\emph{Reacher.}
Because the public DINO-WM release does not include Reacher, we adopt the environment, dataset, and evaluation protocol of LeWorldModel~\cite{maes2026leworldmodel}: 10{,}000 episodes of 201 frames, goals sampled 25 steps ahead within recorded episodes, an interaction budget of 50 environment steps, and success defined as bringing both joints within 0.05 radians of the target.
Planning uses CEM with 300 samples, 30 elites, and 30 iterations over a 5-step horizon.
We use LeWorldModel's frozen-encoder baseline architecture, a 6-layer causal transformer predictor without proprioceptive inputs, trained under our unified schedule for 26{,}000 steps (roughly four passes over the training clips, compared with ten in the default schedule) using mixed precision and unscaled gradient clipping at 1.0.
A random policy succeeds on 6\% of the 50 evaluation episodes.
\Cref{tab:planning} reports mean success across three planner seeds (42, 43, and 44) on the same 50 episodes, with per-seed results listed in \cref{tab:reacher_seeds}.
As in the DINO-WM tasks, every encoder outputs 196 tokens (with DINOv2 taking $196\times196$ inputs); following the LeWorldModel baseline, tokens on Reacher are additionally $\ell_2$-normalized prior to per-channel standardization (computed over 4{,}096 training frames), and encoder inference runs in float32.

\input{tab/reacher_suppl}

\emph{Prediction accuracy.}
We also evaluate open-loop state prediction accuracy for the trained world models of \cref{tab:planning} across PushT, Wall, Granular, Rope, and Reacher, directly testing learned latent dynamics rather than the upstream encoder tokens probed in \cref{tab:main}.
For each encoder and task, we train a state probe on real frozen tokens from the training set and freeze it; given a single frame's tokens, the probe estimates the block pose on PushT, the agent position on Wall, 256 particle coordinates on Granular and Rope, or the two joint angles on Reacher (\cref{tab:wmprobe}).
We then apply this frozen probe to tokens predicted 1 to 5 world-model steps ahead from real history frames and recorded actions (corresponding to 5--25 simulator steps on PushT, Wall, and Reacher with frame skip 5, and 1--5 steps on Rope and Granular without frame skip) across 1{,}500--2{,}000 held-out windows (100 newly collected episodes on Reacher).
\Cref{tab:predacc} in the main text reports the 5-step prediction error, while \cref{tab:predacc_full} provides errors at all horizons alongside two baselines: probing the true future tokens at step 5 (which measures the probe's readout ceiling and serves as a reference rather than a strict lower bound) and copying the last observed frame's tokens.
Across tasks, \ours{} achieves lower 5-step prediction error than both SigLIP~2 and our first-stage encoder on PushT, Granular, Rope, and Reacher, and outperforms DINOv3 on PushT while matching it on Granular and Rope, whereas C-RADIOv3 and DINOv3 attain lower error on Reacher.
Thus, tokens that yield the lowest single-frame readout error are not always the most predictable under learned dynamics.
On Wall, probing true future tokens 5 steps ahead already incurs an error of 9.7--9.8 (matching the error on predicted tokens), so this task does not meaningfully separate the encoders.

\emph{Inputs to the world model and state probe.}
On the four DINO-WM tasks, each history token concatenates the 768 standardized visual channels from the frozen encoder with a 10-dimensional proprioceptive embedding and an action embedding; on Reacher, tokens concatenate only the action embedding (\cref{tab:wmprobe}).
During open-loop rollouts, each predicted frame is fed back as history along with its predicted proprioceptive channels on the DINO-WM tasks, while the action channels are populated with the ground-truth recorded actions (as analyzed in \cref{tab:wmdiag}).
The state probe itself reads only the 768 visual channels of the 196 tokens from a single frame (excluding all proprioceptive and action channels), so both real and predicted visual tokens reside in the same standardized space and are evaluated by the exact same probe.
Because this token space depends only on the frozen encoder and its training-set statistics, a single trained probe per encoder and task evaluates all world-model variants of that encoder.

\emph{Probe training.}
Each task probe is trained exclusively on training-split frames, with checkpoint selection based on validation loss over 2{,}000 held-out training frames; the evaluation trajectories remain strictly unseen until testing.
It shares the width, number of queries, depth, optimizer, learning rate, schedule, training steps, and batch size of the LIBERO probe above, while differing in its single-frame action-free input, 8-head blocks without query self-attention, task-specific losses, weight decay, validation interval, and omission of gradient clipping (\cref{tab:wmprobe}).

\input{tab/wmprobe_suppl}
\input{tab/decode_suppl}

\emph{Decoded frames.}
We evaluate single-frame reconstruction on PushT following the adapter protocol of \cref{tab:main}.
For each encoder, an evaluation decoder is trained on 19{,}000 frames randomly sampled from the training trajectories and evaluated on 1{,}000 frames sampled uniformly across the 21 validation trajectories (\cref{tab:decode}).
All encoders achieve $\ell_1$ errors between .0055 and .0066, compared with .0044 for the standalone SDXL autoencoder and .0370 for predicting the mean training frame.
Our first-stage encoder achieves the lowest error (.0055), while \ours{} (.0061) performs comparably to the public encoders (.0062--.0066).

\emph{Rendered predictions.}
We use these same evaluation decoders to render the future tokens predicted by the world models into pixel space.
For each predicted token, we retain the 768 visual channels (discarding proprioceptive and action channels) and invert the world model's per-channel standardization via $z\,\sigma + \mu$; real tokens undergo the identical inverse transformation before decoding.
Because each decoder was trained on raw frozen tokens from the 19{,}000 training frames at $224\times224$ resolution, the unstandardized world-model tokens on the last observed frame match the decoder's expected input exactly (cosine similarity 1.000) for all encoders except DINOv2, whose world model operates at $196\times196$ resolution (cosine similarity .80) and is therefore omitted from pixel rendering.
We decode tokens predicted 1 to 5 world-model steps ahead across a fixed random sample of 200 validation windows (out of the 1{,}695 windows in \cref{tab:predacc} spanning all 21 validation trajectories) and compare the rendered images against ground-truth future frames (\cref{tab:render}).
All 200 windows come from held-out validation trajectories unseen during both world-model and decoder training.
At 5 world-model steps ahead, rendered predictions from \ours{} achieve an LPIPS of .020, second only to C-RADIOv3 (.018) and outperforming DINOv3 (.024), our first-stage encoder (.025), and SigLIP~2 (.031).

\input{tab/render}

\begin{figure*}[p]
    \centering
    \includegraphics[width=\textwidth,page=1]{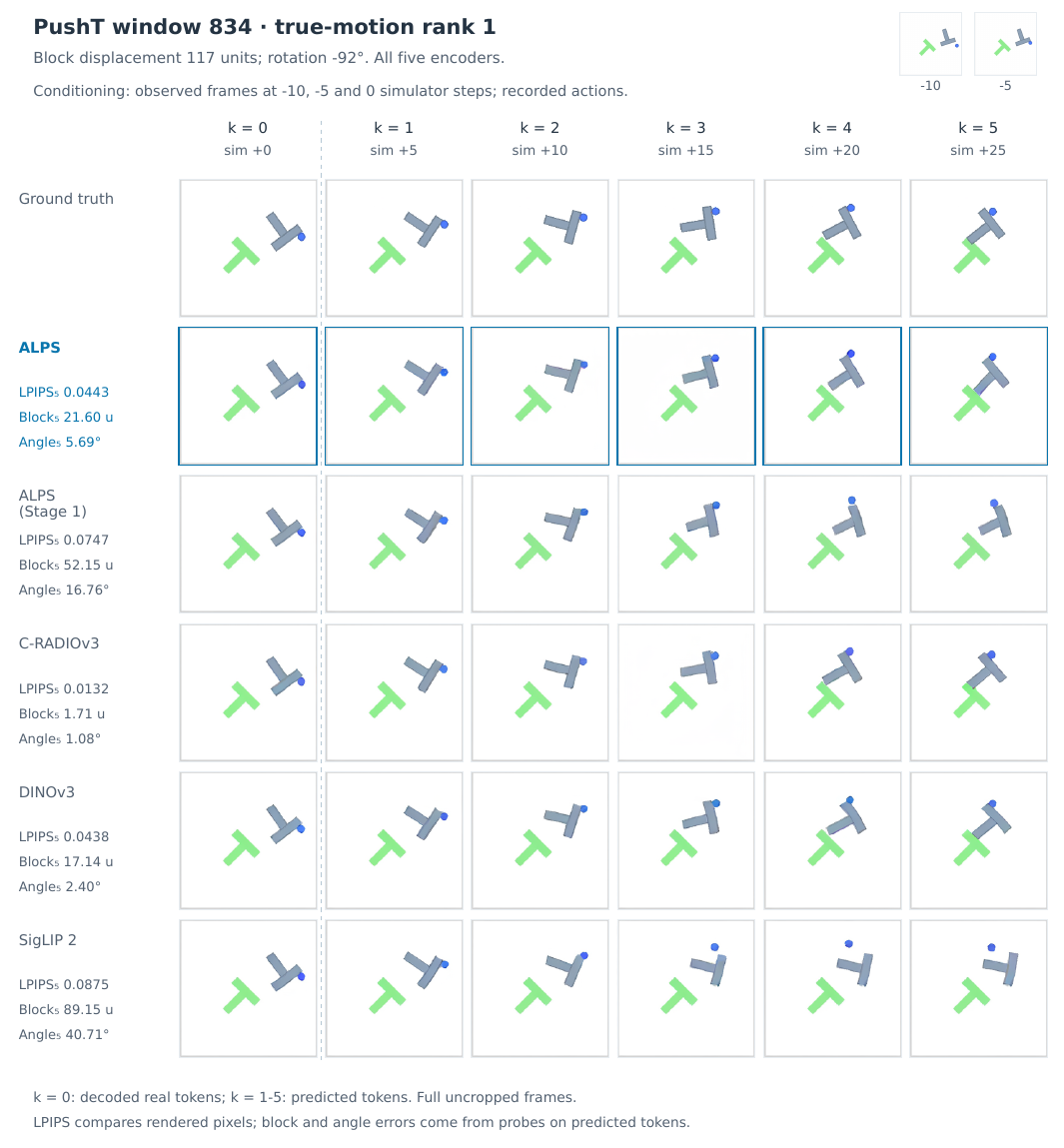}
    \caption{\textbf{Rendered world-model predictions across encoders on PushT window 834}, ranked first by our motion criterion (\cref{sec:suppl_render}).
Each row applies an encoder's evaluation decoder to its real tokens at the current frame ($k = 0$) and to tokens predicted $k = 1$ to 5 world-model steps ahead; the top row shows ground-truth frames.
Row annotations report 5-step LPIPS against the true frame alongside block position and angle errors read by the probe of \cref{tab:predacc}.}
    \label{fig:render_w834}
\end{figure*}

\begin{figure*}[p]
    \centering
    \includegraphics[width=\textwidth,page=2]{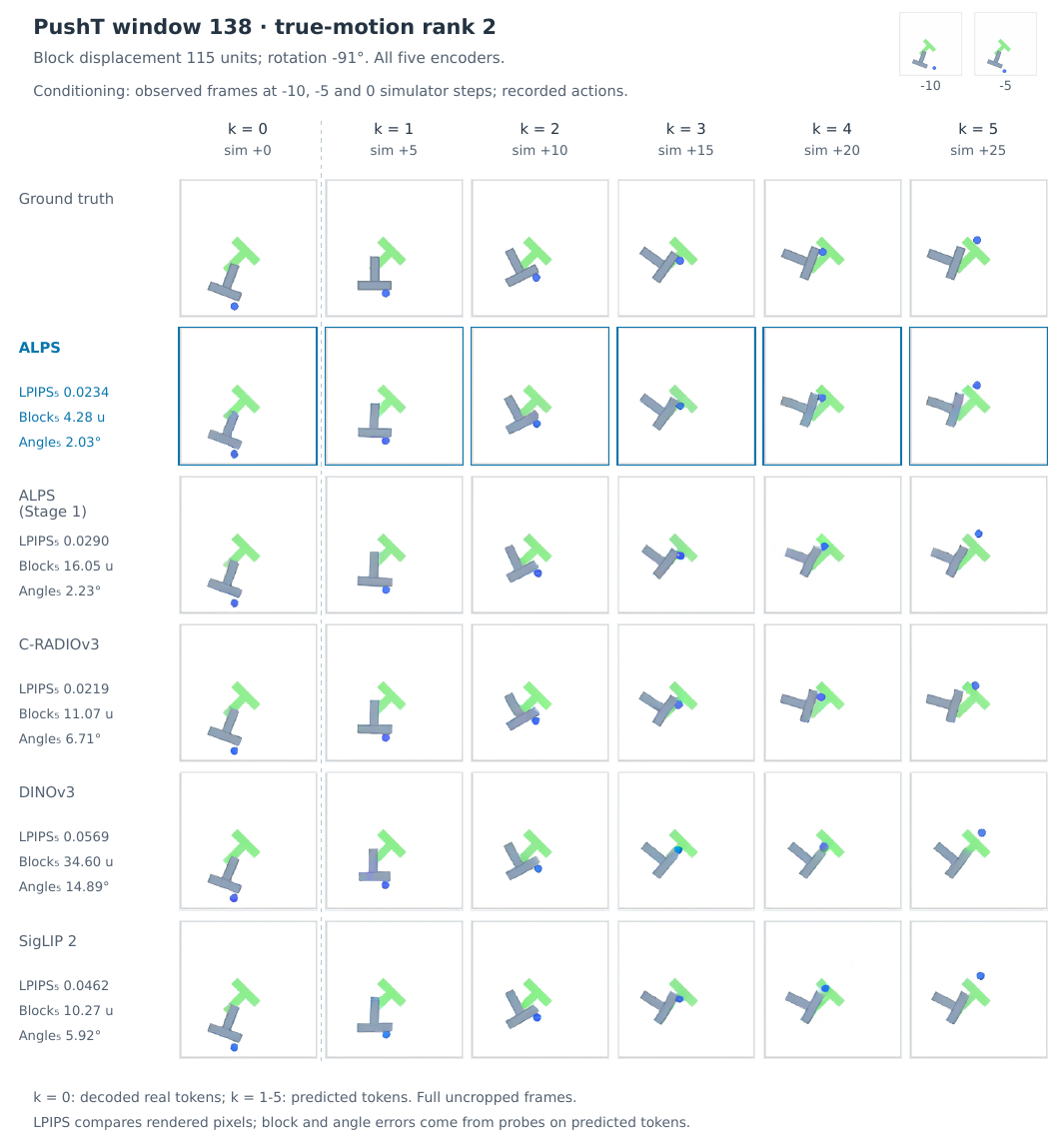}
    \caption{\textbf{Rendered world-model predictions across encoders on PushT window 138}, ranked second by our motion criterion; layout and metrics follow \cref{fig:render_w834}.}
    \label{fig:render_w138}
\end{figure*}

\begin{figure*}[p]
    \centering
    \includegraphics[width=\textwidth,page=3]{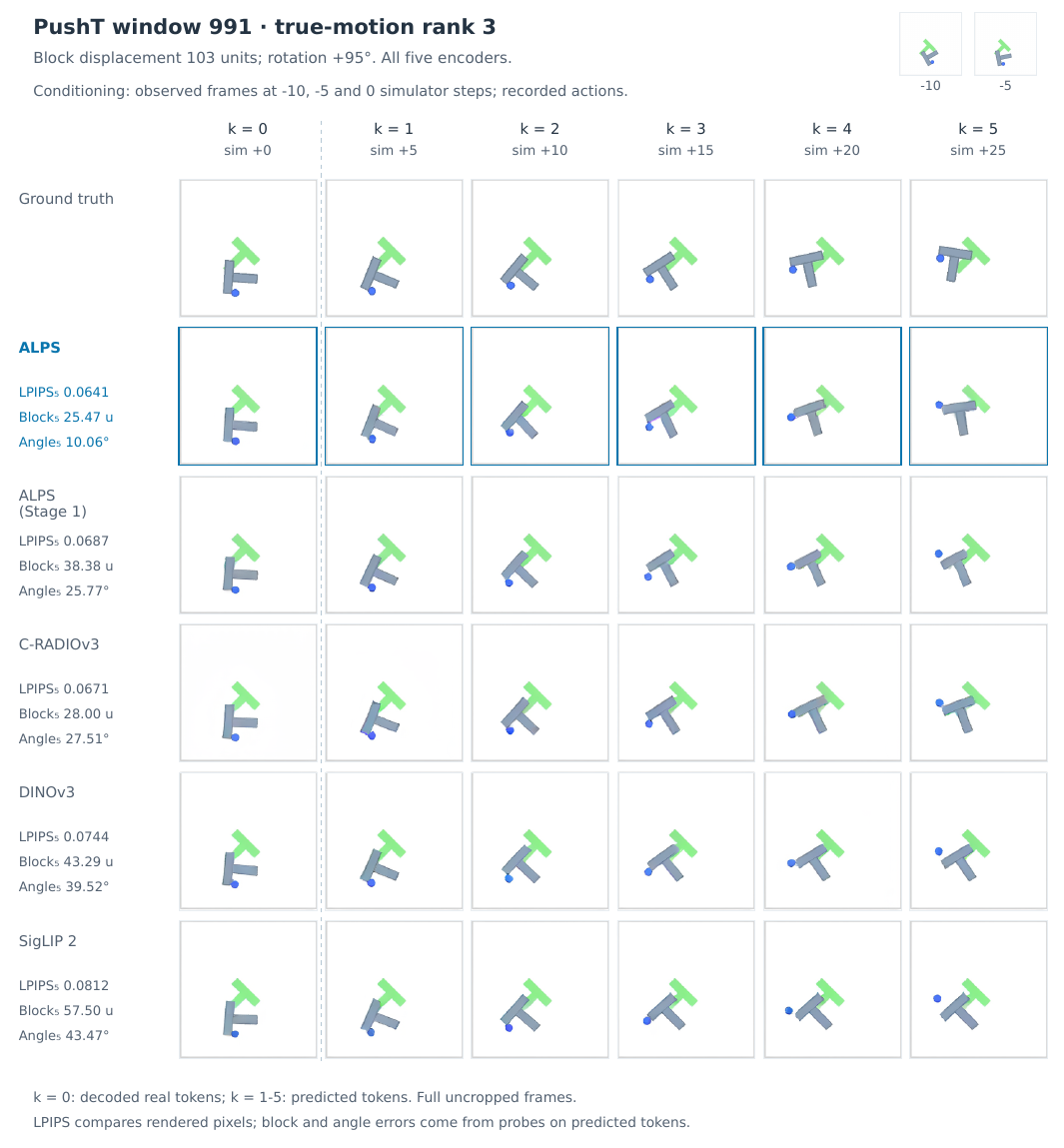}
    \caption{\textbf{Rendered world-model predictions across encoders on PushT window 991}, ranked third by our motion criterion; layout and metrics follow \cref{fig:render_w834}.}
    \label{fig:render_w991}
\end{figure*}

\emph{Rendered predictions on the windows with the largest motion.}
\label{sec:suppl_render}
Most of the 200 validation windows involve modest object motion (median block displacement of 31 units over 5 steps).
To visualize predictions under substantial motion without cherry-picking, we ranked all 200 windows by ground-truth block motion using a rule fixed prior to visual inspection: we filtered for windows with at least 40 units of displacement or 20 degrees of rotation over the 5 steps, and ranked them by displacement plus twice the absolute rotation in degrees.
The top three windows are 834 (117 units, $-92^\circ$), 138 (115 units, $-91^\circ$), and 991 (103 units, $+95^\circ$); \cref{fig:render_main} in the main text displays window 834 for \ours{} and C-RADIOv3, while \cref{fig:render_w834,fig:render_w138,fig:render_w991} show all three windows across all rendered encoders.
Evaluated by the state probe of \cref{tab:predacc} on 5-step predicted tokens, block position error on window 834 ranges from 17 to 89 units across all encoders except C-RADIOv3 (2 units), with \ours{} at 22 units (up from 3 units at step 1) and under $6^\circ$ of orientation error; on window 138, error ranges from 4 units (\ours{}) to 35 units (DINOv3); and on window 991, it ranges from 25 units (\ours{}) to 58 units (SigLIP~2).

\input{tab/predacc_full_suppl}
\input{tab/wmdiag}

\emph{Readout, one-step prediction, and rollout.}
\label{sec:suppl_wmdiag}
To disentangle token readout error, single-step dynamics error, and multi-step compounding error, we evaluate the state probe of \cref{tab:predacc} under four conditions across the 200 validation windows of \cref{tab:render} using the final world-model checkpoints from \cref{tab:planning} (\cref{tab:wmdiag}).
\emph{True future tokens} apply the probe directly to the frozen encoder's tokens of the ground-truth frame $k$ steps ahead.
\emph{One-step prediction from true history} uses teacher forcing: at each step $k$, the predictor receives the ground-truth tokens (including proprioception) of the three preceding frames along with recorded actions to predict step $k$, never conditioning on its own prior predictions.
The \emph{five-step open-loop rollout} matches the protocol of \cref{tab:predacc} and \cref{fig:render_main}: starting from three observed frames, each predicted frame and its predicted proprioception are fed back autoregressively as history for the next step, using only the recorded actions.
Two additional baselines complete the table: an open-loop rollout where fed-back proprioceptive channels are replaced with ground-truth recorded proprioception, and a static baseline that repeats the last observed frame, confirming that the world models capture nontrivial motion.
At $k = 5$, the error gap between \ours{} and C-RADIOv3 remains nearly constant across settings: 0.7 units on true future tokens, 0.7 on one-step predictions, and 0.8 after the five-step rollout (whose confidence interval includes zero). Notably, autoregressive rollout adds only about 3.0 units of error over one-step prediction for both \ours{} and C-RADIOv3, compared with 5.7 to 8.9 units for our first-stage encoder, DINOv3, and SigLIP~2.
Rendered pixel predictions follow the same trend across the three conditions (LPIPS of .0068 vs.\ .0070, .0083 vs.\ .0077, and .0203 vs.\ .0176).
Replacing predicted proprioception with ground-truth proprioception during rollout reduces error for C-RADIOv3 but not for \ours{}, widening their gap to 1.6 units.
Across the full 1{,}695 windows of \cref{tab:predacc}, the post-rollout gap is larger than the true-token readout gap (1.1 vs.\ 0.2 units; \cref{tab:predacc_full}), indicating that the fraction of error attributable to rollout accumulation varies across trajectory samples.

\emph{Pipeline calibration.}
\label{sec:suppl_wmold}
To verify our implementation against prior work, we trained DINOv2-S/14 (the default encoder in DINO-WM) using DINO-WM's original recipe and training budget; it achieves 94\% success on PushT and 92\% on Wall across 50 goals, closely matching both the numbers reported in DINO-WM (90\% and 96\%) and our evaluation of their released checkpoints (92\% and 94\%).

\begin{figure*}[p]
    \centering
    \includegraphics[width=0.85\textwidth]{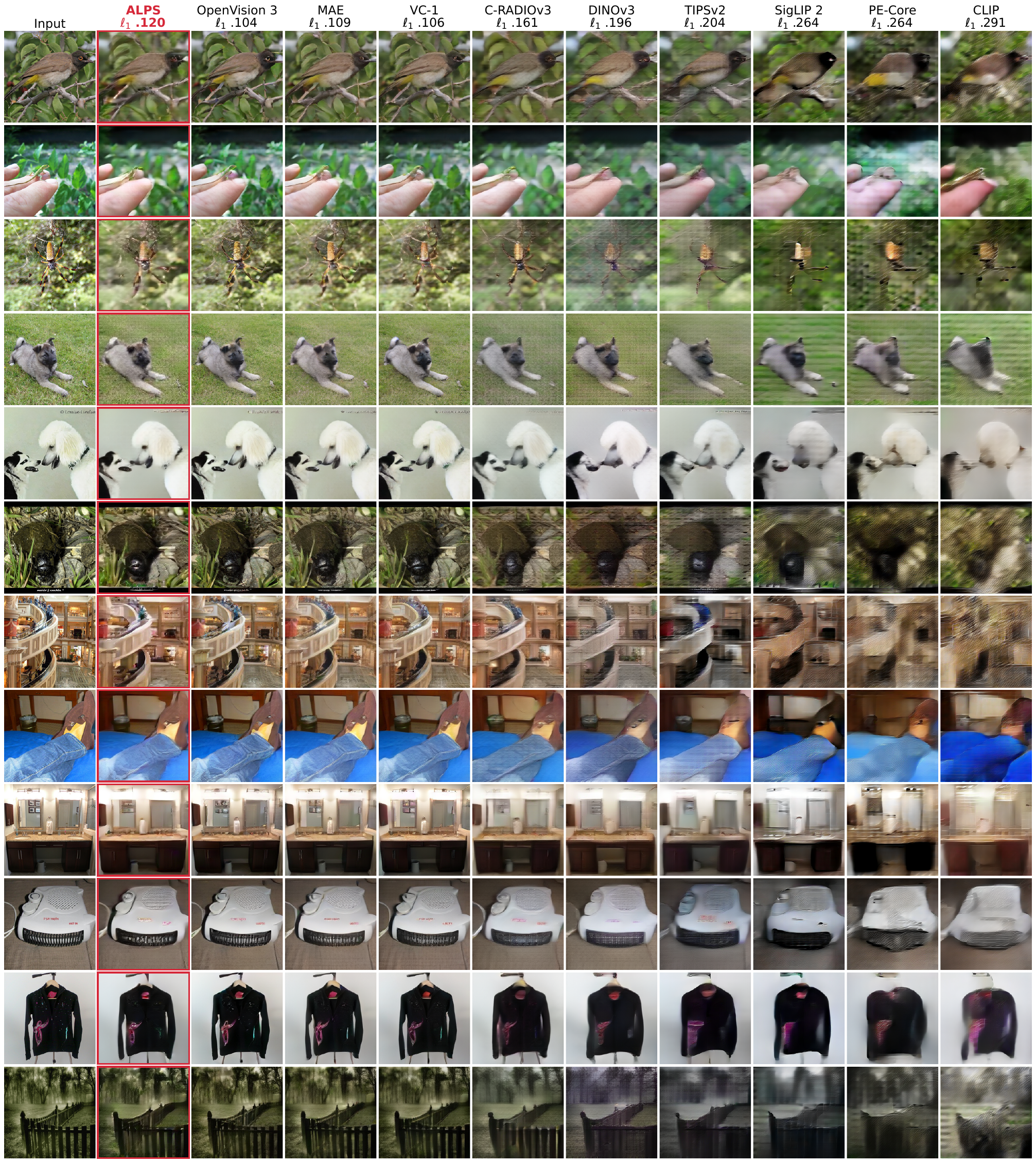}
    \caption{\textbf{Reconstruction from frozen tokens on a random sample of held-out ImageNet images}: input images alongside outputs from the evaluation decoders of ten encoders, each labeled with its mean $\ell_1$ error over the 1{,}000 held-out images (the first three rows correspond to \cref{fig:recon}).}
    \label{fig:recon_all}
\end{figure*}

\paragraph{Reconstruction.}
The evaluation decoder, trained from scratch on frozen tokens for each encoder and separate from our \stageone{} pretraining decoder, maps patch tokens to the latent space of a frozen SDXL autoencoder~\cite{podell2024sdxl}, whose decoder renders the final image. Specifically, it consists of a $1\times1$ convolution to width 512, a transposed convolution upsampling the $14\times14$ token grid to $28\times28$ (bilinearly resizing token grids that differ from $14\times14$ beforehand), five residual blocks, and a $3\times3$ convolution to the four latent channels.
We train the adapter using AdamW (learning rate $10^{-4}$) for 20k steps at batch size 8, minimizing the sum of latent-space MSE and decoded-image LPIPS while injecting Gaussian noise of random scale into the input tokens. Training uses 19{,}000 images from the ImageNet-1k validation set (the first 19 sorted files per class), and evaluation uses the 20th file per class (1{,}000 held-out images), with all images resized to a short side of 224 pixels and center-cropped.
We compute $\ell_1$ reconstruction error on pixel values scaled to $[-1, 1]$.
Passing the held-out images directly through the SDXL autoencoder without token compression yields an $\ell_1$ error of .071 and an LPIPS of .130 (serving as an upper-bound reference), whereas predicting the mean training image yields .460 and .816.
\Cref{fig:recon_all} displays reconstructions on a random sample of twelve held-out images across ten encoders, the first three of which appear in \cref{fig:recon} of the main text.
For V-JEPA~2, each static image is duplicated into a two-frame clip.

\input{tab/recon_large}

\emph{Stronger decoding and training configuration.}
As a capacity control, we re-evaluate four encoders using a larger adapter (width 1{,}024 and ten residual blocks) trained for 60k steps on 49{,}000 ImageNet validation images (all 49 remaining files per class, keeping the 20th file strictly held out) and tested on the same 1{,}000 held-out images (\cref{tab:recon_large}).
Because this setting increases adapter width, depth, training steps, and training set size simultaneously, it tests an overall stronger decoding regime rather than model capacity in isolation.
All four encoders improve on both metrics.
On $\ell_1$ error, \ours{} improves from .120 to .107 and C-RADIOv3 from .161 to .141, maintaining a consistent relative advantage for \ours{} (25\% lower error with the standard adapter and 24\% lower with the stronger configuration), while DINOv2 and SigLIP~2 remain at roughly twice our error (.218 and .220).
On LPIPS, \ours{} moves from 4\% behind C-RADIOv3 under the standard adapter (.372 vs.\ .359) to 3\% ahead under the stronger configuration (.302 vs.\ .311), while DINOv2 and SigLIP~2 reach .423 and .404.

\paragraph{Temporal transfer.}
\label{sec:suppl_video}

\input{tab/video}
\input{tab/stage2_full}

Because \ours{} is pretrained exclusively on static images and text, we evaluate how its spatial representations transfer to video across two complementary benchmarks (\cref{tab:video}): video object segmentation on DAVIS-2017~\cite{pont20172017} via training-free mask propagation ($\mathcal{J}\&\mathcal{F}$), which tests cross-frame spatial correspondence, and action recognition on Something-Something~v2 (SSv2)~\cite{goyal2017something} using an attentive probe trained for 20 epochs on 16-frame clips at $224\times224$ resolution, which tests fine-grained temporal dynamics.
\stagetwo{} refinement improves DAVIS $\mathcal{J}\&\mathcal{F}$ from .456 to .533 while leaving SSv2 largely unchanged (56.4 vs.\ 55.6); static cross-view refinement thus sharpens frame-to-frame spatial correspondence without supplying fine-grained motion features, underscoring why physical-state representation and action classification should be evaluated separately.
Extending cross-view refinement to video clips and interaction trajectories is a natural next step (\cref{sec:conclusion}).

On DAVIS validation videos, we follow the training-free label propagation protocol of DINO~\cite{caron2021emerging}: $\ell_2$-normalized last-layer patch tokens in each frame are compared against those of the first frame and the preceding seven frames within a 12-token spatial radius, propagating the soft masks of the top five neighbors (temperature 0.1).
Video frames are resized to a short side of 480 pixels (476 for patch size 14), with positional embeddings interpolated to the resulting token grid; for our encoders, which use fixed sine-cosine embeddings, spatial coordinates are rescaled to match the $14\times14$ pretraining range.
V-JEPA~2 processes 16-frame clips up to the current frame (padded at the start by repeating the first frame), from which we extract the final tubelet's tokens and rescale spatial rotary positions to its $16\times16$ pretraining grid in the same manner.
Without spatial position rescaling, its rotary embeddings extrapolate poorly to 480p inputs, dropping its $\mathcal{J}\&\mathcal{F}$ score from .526 to .347.
On Something-Something~v2, V-JEPA~2 reaches 70.3\% under the same position rescaling and probe protocol (re-initializing the input projection for its 1{,}024-dimensional tokens and matching temporal embeddings to its 8 tubelets); because SSv2 is part of V-JEPA~2's pretraining corpus, this score serves as an in-domain reference rather than an out-of-distribution transfer result.

\paragraph{Precision.}
Video object segmentation (DAVIS), action recognition (SSv2), and policy learning (\cref{tab:policy_hparams}) run frozen encoder inference in bfloat16; running float32 inference on DAVIS shifts $\mathcal{J}\&\mathcal{F}$ slightly from .456 and .533 to .462 and .535 (\cref{tab:video}).
Language evaluations run public encoders with float16 autocast and our encoders in float32 (since our encoders output unnormalized patch tokens); for SigLIP~2, switching to float32 changes zero-shot accuracy and retrieval by at most 0.06 points.

\section{\stagetwo{}: full comparison and image source}
\label{sec:suppl_variants}

\Cref{tab:stage2_full} reports the complete state-probe and language-alignment metrics for the refinement comparison in \cref{fig:refine} under a shared budget of 5M sampled images.
For our encoder, \stagetwo{} boosts LIBERO-Spatial policy success by 9.7 points while preserving performance on Goal ($+0.8$ points) and Long ($+0.8$ points); for SigLIP~2, refinement shifts policy success by at most 2.1 points across the three suites.
When refining SigLIP~2 and CLIP, both the student and frozen teacher are initialized from the released image tower, prediction targets are taken after the final normalization layer (matching downstream evaluation) and standardized over 32{,}768 training images, and the language projection head remains frozen.
Downstream, refined SigLIP~2 reaches 79.0\% on LIBERO-Goal and 82.4\% on Spatial (compared with 85.1\% and 86.5\% for \ours{}), and 74.4\% on Long (compared with 69.7\%).
Replacing our curated high-resolution image collection in \stagetwo{} with single web images from DataComp~\cite{gadre2023datacomp} yields a state-probe error of .886 (vs.\ .829), DAVIS $\mathcal{J}\&\mathcal{F}$ of .523 (vs.\ .533), and ImageNet zero-shot accuracy of 73.6 (vs.\ 73.4); because DataComp images have substantially lower native resolution (39\% have a short side below 224 pixels, compared with 512--1{,}024 pixels for most images in our curated set), this gap reflects both scene composition and source resolution.

\section{Details of the first-stage ablation}
\label{sec:suppl_ablation}

Each ablation variant in \cref{tab:ablation} is pretrained from scratch for 10\% of the \stageone{} schedule (39k steps at $128\times128$ resolution followed by 6.3k steps at $224\times224$) and subsequently refined with the full \stagetwo{} recipe (4.9k steps, 5M sampled images); variants without the captioning loss train the text tower solely via contrastive alignment.
Text-to-image retrieval mirrors the image-to-text trends: after \stageone{}, recall is highest for the contrastive-only variant (47.6 on COCO and 74.4 on Flickr30k) and lowest when removing captioning (45.9 and 72.5), while \stagetwo{} reduces recall by only 0.4--0.8 points on COCO and 1.1--1.5 points on Flickr30k across all variants.
Because all variants retain contrastive alignment, the ablation isolates the three auxiliary objectives at one-tenth pretraining scale: removing captioning or reconstruction lowers state-probe error (by 4.3\% and 8.5\% after \stageone{}, and by 10.4\% and 13.4\% relative to the full recipe after \stagetwo{}), removing dense correspondence changes state-probe error by less than 1\% but degrades cross-view patch matching (\cref{tab:dense_match}), and none of the auxiliary objectives is required to preserve language alignment during \stagetwo{}.

\emph{Cross-view patch matching.}
To directly evaluate the correspondence supervised by $\mathcal{L}_{\text{dense}}$, we take the 1{,}000 held-out ImageNet images from our reconstruction benchmark and extract four pairs of square crops from each image's central square, resizing every crop to $224\times224$: (1) the full central square paired with a crop covering 50\%--100\% of its area, randomly flipped with probability 0.5 (matching the view distribution of $\mathcal{L}_{\text{dense}}$); (2) the same setup using smaller crops of 10\%--50\% area; (3) the first setting with \stagetwo{} photometric augmentations applied to the crop; and (4) two partially overlapping crops each covering 30\%--70\% of the central square.
Each patch in the second view whose center falls inside the first view is matched to its nearest neighbor in the first view by cosine similarity of last-layer tokens, and a match is scored as correct if it selects the patch containing the true geometric center.
\Cref{tab:dense_match} reports mean matching accuracy across the four view settings: removing $\mathcal{L}_{\text{dense}}$ reduces accuracy by 6.0 points after \stageone{} (37.4\% to 31.4\%) and by 2.6 points after \stagetwo{} (40.8\% to 38.2\%).

\begin{table}[t]
\centering
\footnotesize
\setlength{\tabcolsep}{6pt}
\begin{tabular}{l cc}
\toprule
 & \multicolumn{2}{c}{Patch matching (\%)\,$\uparrow$} \\
\cmidrule(lr){2-3}
\stageone{} objectives & \stageone{} & \stagetwo{} \\
\midrule
all four objectives     & \textbf{37.4} & 40.8 \\
no reconstruction       & 34.6 & 38.6 \\
no dense correspondence & 31.4 & 38.2 \\
no captioning           & 37.0 & \textbf{41.5} \\
contrastive only        & 29.7 & 35.1 \\
\bottomrule
\end{tabular}
\caption{\textbf{Cross-view patch matching for the ablation variants of \cref{tab:ablation}}, evaluated after \stageone{} and after \stagetwo{}.
Accuracy measures the percentage of patches whose nearest token neighbor in a second view is the geometrically corresponding patch, averaged across four view-pair settings on 1{,}000 held-out ImageNet images.}
\label{tab:dense_match}
\end{table}

\section{Representation drift under \stagetwo{}}
\label{sec:suppl_drift}

We pass all 5{,}000 COCO test images through each encoder before and after \stagetwo{} refinement (using each model's language-evaluation preprocessing) and measure feature drift both on corresponding last-layer patch tokens and on the global image embeddings produced by the frozen language head (the projected mean token for \ours{}, attention pooling for SigLIP~2, and the projected class token for CLIP).
For standardized token comparisons, each channel is standardized using the pre-refinement statistics of the respective encoder.
For our encoder (\stageone{} vs.\ \ours{}), mean cosine similarity before and after refinement is .88 on raw tokens, .84 on standardized tokens, and .96 on global language embeddings, with a linear centered kernel alignment (CKA)~\cite{kornblith2019similarity} of .83 on patch tokens; by comparison, SigLIP~2 reaches .81, .72, and .84 (linear CKA .74), and CLIP reaches .77, .71, and .83 (linear CKA .59).
Our representation thus drifts less than those of SigLIP~2 and CLIP (particularly in its global language embedding), and as shown in \cref{sec:ablation}, our linear mean-pooled language interface preserves alignment through refinement even without any single auxiliary objective in \stageone{}.

\emph{SigLIP~2 with a smaller encoder learning rate.}
We repeat \stagetwo{} refinement on SigLIP~2 with the encoder learning rate reduced from $10^{-5}$ to $3 \times 10^{-6}$ and all other hyperparameters unchanged.
Its state-probe error improves from 1.160 to .959 and its DAVIS score from .503 to .553, while ImageNet zero-shot accuracy still drops by 1.9 points, image-to-text Recall@1 drops by 2.3 points on Flickr30k and 2.9 on COCO, and text-to-image Recall@1 drops by 1.0 and 0.6 points.
On the COCO test set, this lower-learning-rate variant yields a mean cosine similarity of .88 on raw tokens, .81 on standardized tokens, and .97 on global language embeddings, with a linear CKA of .81.
Notably, although this SigLIP~2 variant retains slightly higher mean image-embedding similarity than \ours{} (.97 vs.\ .96), it still suffers a larger drop in retrieval accuracy.
Maintaining high average cosine similarity is therefore insufficient on its own to prevent language degradation under nonlinear attention pooling or class-token readouts.
Instead, these results suggest that pairing unnormalized patch tokens with a linear mean-pooled language readout provides a more stable interface under local cross-view refinement.

\clearpage

%% file: tab/stage1_hparams.tex
\begin{table*}[t]
\centering
\small
\setlength{\tabcolsep}{4pt}
\begin{tabular}{ll}
\toprule
\multicolumn{2}{l}{\emph{Architecture}} \\
encoder & ViT-B/16, width 768, 12 layers, 12 heads, MLP 3{,}072 \\
 & fixed 2D sine-cosine position embedding, no final normalization \\
 & one class token, used in attention but not in any loss \\
image embedding & mean-pooled patch tokens, linear projection to 768 \\
text encoder & 12 layers, width 768, 32 tokens, pooled at the last token \\
caption decoder & 12 layers, width 768, prefix over image tokens, 80 text tokens \\
tokenizer & WordPiece, BERT vocabulary (lowercase) \\
pretraining decoder & $1 \times 1$ conv to 512 channels, 4 blocks with 256, 128, 64, 32 \\
 & channels ($3 \times 3$ conv, pixel shuffle $\times 2$, GroupNorm, SiLU, $3 \times 3$ conv), \\
 & $3 \times 3$ conv to RGB; 7.4M parameters \\
dense head & linear 768 to 256 with normalization, shared by both views \\
\midrule
\multicolumn{2}{l}{\emph{Views}} \\
input $x$ & Inception crop, area 0.4 to 1, no color augmentation \\
dense view $g(x)$ & square crop of $x$, area 0.5 to 1, horizontal flip with $p = 0.5$ \\
\midrule
\multicolumn{2}{l}{\emph{Optimization}} \\
optimizer & AdamW, $\beta_1 = 0.9$, $\beta_2 = 0.95$, no gradient clipping \\
weight decay & 0.05 on weight matrices, none on biases, norms, embeddings \\
schedule & linear warm-up, then cosine decay to 0 \\
precision & bfloat16 compute, 32-bit parameters \\
\midrule
\multicolumn{2}{l}{\emph{Phases}} \\
128 pixels & batch 32{,}768, peak lr $1.0 \times 10^{-3}$, 390{,}981 steps, warm-up 1\% \\
224 pixels & batch 16{,}384, peak lr $2.6 \times 10^{-5}$, 62{,}556 steps, warm-up 5\% \\
 & initialized from the 128-pixel weights with a fresh optimizer state \\
\bottomrule
\end{tabular}
\caption{\textbf{Settings of \stageone{}.} Loss weights and objectives: \cref{sec:stage1,sec:suppl_train}.}
\label{tab:stage1_hparams}
\end{table*}

%% file: tab/stage2_config_suppl.tex
\begin{table*}[t]
\centering
\small
\setlength{\tabcolsep}{4pt}
\begin{tabular}{ll}
\toprule
\multicolumn{2}{l}{\emph{Views}} \\
crop frame & central square of the image, side $m$; square boxes of side $\sqrt{a}\,m$ in integer pixels, sampled uniformly \\
teacher crop $B$ & area $a \sim \mathcal{U}(0.4, 1)$; clean; resized to $224 \times 224$ pixels (bilinear) \\
student crop $A$ & local, $a \sim \mathcal{U}(0.1, 0.4)$, or global, $a \sim \mathcal{U}(0.4, 1)$, $p = 0.5$; resized to $224 \times 224$ pixels (bilinear) \\
 & photometric augmentation; at least 16 patch centers of $B$ inside $A$, else up to 10 resamples, then $A = B$ \\
token grids & $14 \times 14$ patches of 16 pixels for both crops; no flip \\
\midrule
\multicolumn{2}{l}{\emph{Predictor $P$}} \\
queries & one per patch of $B$ (196); learned vector, position code, and learned linear map of $(\ln r, (\ln r)^2)$ \\
position & center of each patch of $B$ in the frame of $A$ (in patch units of $A$); patch $(u, v)$ is centered at $(u + 0.5, v + 0.5)$ \\
relative scale & $r = s_B / s_A$ (side of $B$ over side of $A$); natural logarithm \\
position code & fixed 2D sine-cosine code, 128 frequencies $10000^{-k/128}$ per axis; also added to the student tokens \\
keys and values & student tokens $E(A)$ projected to width 512, plus the position code of their patch centers and one constant register token \\
layer & self-attention among all 196 queries, cross-attention from queries to keys, MLP of width 2{,}048; pre-norm \\
size & 4 layers, width 512, 8 heads; output: normalization and linear map to 768, then normalization and linear map \\
\midrule
\multicolumn{2}{l}{\emph{Targets}} \\
teacher & frozen \stageone{} encoder on $B$; tokens standardized per channel, $\bar z = (z - \mu) / \sigma$ \\
$\mu$, $\sigma$ & mean and standard deviation of each of the 768 channels over all 196 positions of 32{,}768 training images, computed once \\
 & (square crops of 80\% to 100\% of the central square at 224 pixels, no augmentation, \stageone{} encoder) \\
loss & squared error averaged over the 768 channels and patches of $B$ centered inside $A$, then across images \\
\bottomrule
\end{tabular}
\caption{\textbf{Settings of \stagetwo{} for \ours{}.} Optimization and augmentation details: \cref{sec:suppl_train}.}
\label{tab:stage2_config}
\end{table*}

%% file: tab/stage2_sources.tex
\begin{table*}[t]
\centering
\small
\setlength{\tabcolsep}{3pt}
\begin{tabular}{lr lr}
\toprule
source & share (\%) & source & share (\%) \\
\midrule
SEED-Data-Edit~\cite{ge2024seed} & 17.5 & PICA-100K~\cite{pu2025picabench} & 1.8 \\
ScaleEdit-12M~\cite{chen2026scaleedit} & 14.9 & NeIn~\cite{bui2025nein} & 1.2 \\
UltraEdit~\cite{zhao2024ultraedit} & 10.1 & Pico-Banana~\cite{qian2026pico} & 1.1 \\
PIPE~\cite{wasserman2025paint} & 9.7 & Dress-ED~\cite{lobba2026dress} & 1.0 \\
AnyEdit~\cite{yu2025anyedit} & 9.5 & WEAVE~\cite{chow2026weave} & 0.7 \\
OmniEdit~\cite{wei2025omniedit} & 8.4 & GPT-Image-Edit~\cite{wang2025gpt} & 0.3 \\
NHR-Edit~\cite{kuprashevich2026nohumansrequired} & 6.0 & InstructPix2Pix~\cite{brooks2023instructpix2pix} & 0.2 \\
CrispEdit-2M~\cite{chow2026editmgt} & 5.7 & RefEdit~\cite{pathiraja2025refedit} & 0.1 \\
UnicEdit-10M~\cite{ye2026unicedit} & 5.1 & HumanEdit~\cite{bai2024humanedit} & 0.1 \\
ImgEdit~\cite{ye2026imgedit} & 4.2 & HQ-Edit~\cite{hui2024hq} & 0.0 \\
AURORA (with MagicBrush)~\cite{krojer2024learning,zhang2023magicbrush} & 2.5 & \textbf{total} & \textbf{100} \\
\bottomrule
\end{tabular}
\caption{\textbf{Sources of the \stagetwo{} training images} and their share of the 5M sampled images.}
\label{tab:stage2_sources}
\end{table*}

%% file: tab/libero.tex
\begin{table*}[t]
\centering
\small
\setlength{\tabcolsep}{4.2pt}
\fitwidth{\textwidth}{%
\begin{tabular}{l ccc ccc ccc}
\toprule
 & \multicolumn{3}{c}{LIBERO-Object (cm\,$\downarrow$)} & \multicolumn{3}{c}{LIBERO-Spatial (cm\,$\downarrow$)} & \multicolumn{3}{c}{All six cells} \\
\cmidrule(lr){2-4} \cmidrule(lr){5-7} \cmidrule(lr){8-10}
Encoder & readout & +1\,s & +3\,s & readout & +1\,s & +3\,s & mean (cm) & median (cm) & relative \\
\midrule
\multicolumn{10}{l}{\emph{Spatial \& self-supervised encoders}} \\
TIPSv2-B/14~\cite{cao2026tipsv2}                     & .995 & 1.141 & 1.582 & 1.765 & 1.517 & 1.790 & 1.465 & 1.016 & .963 \\
DINOv2-B/14~\cite{oquab2023dinov2}                   & .994 & 1.036 & 1.642 & 1.651 & 1.529 & 1.763 & 1.436 & .938 & .942 \\
PE-Spatial-B/16~\cite{bolya2026perception}           & 1.008 & .989 & 1.340 & 1.944 & 1.574 & 1.827 & 1.447 & .906 & .936 \\
DINOv3-B/16~\cite{simeoni2025dinov3}                 & .910 & 1.041 & 1.184 & 1.592 & 1.557 & 1.738 & 1.337 & .921 & .870 \\
C-RADIOv3-B~\cite{heinrich2025radiov2}               & .907 & .956 & 1.285 & 1.659 & \textbf{1.436} & 1.742 & 1.331 & .893 & .864 \\
\midrule
\multicolumn{10}{l}{\emph{Image-text encoders}} \\
CLIP-B/16~\cite{radford2021learning}                 & 1.092 & 1.151 & 1.884 & 2.137 & 2.093 & 2.949 & 1.884 & 1.148 & 1.194 \\
SigLIP~2-B/16~\cite{tschannen2025siglip}             & 1.130 & 1.207 & 1.487 & 2.286 & 2.333 & 2.499 & 1.824 & 1.225 & 1.160 \\
PE-Core-B/16~\cite{bolya2026perception}              & 1.138 & 1.200 & 1.980 & 2.153 & 1.973 & 2.436 & 1.813 & 1.205 & 1.170 \\
OpenVision-B/16~\cite{li2025openvision}              & 1.168 & 1.095 & 1.530 & 1.809 & 1.819 & 2.022 & 1.574 & 1.074 & 1.027 \\
\midrule
\multicolumn{10}{l}{\emph{Pixel \& latent reconstruction encoders}} \\
MAE-B/16~\cite{he2022masked}                         & .906 & .944 & 1.529 & 1.886 & 2.073 & 2.003 & 1.557 & .886 & .990 \\
VC-1-B~\cite{majumdar2023we}                        & 1.136 & \textbf{.868} & \textbf{1.104} & 1.953 & 1.593 & 2.214 & 1.478 & .906 & .943 \\
OpenVision~3-B~\cite{zhang2026openvision}            & 1.007 & .933 & 1.581 & 2.054 & 1.822 & 2.109 & 1.584 & .993 & 1.011 \\
\midrule
\multicolumn{10}{l}{\emph{Larger encoders, ViT-L (304M to 330M parameters)}} \\
\textcolor{gray}{OpenVision~2-L/14~\cite{liu2026openvision}} & \textcolor{gray}{.996} & \textcolor{gray}{.911} & \textcolor{gray}{1.375} & \textcolor{gray}{1.848} & \textcolor{gray}{1.664} & \textcolor{gray}{2.067} & \textcolor{gray}{1.477} & \textcolor{gray}{1.012} & \textcolor{gray}{.945} \\
\textcolor{gray}{V-JEPA~2-L/16~\cite{assran2025v}}   & \textcolor{gray}{.805} & \textcolor{gray}{.945} & \textcolor{gray}{1.336} & \textcolor{gray}{1.333} & \textcolor{gray}{1.466} & \textcolor{gray}{1.780} & \textcolor{gray}{1.277} & \textcolor{gray}{.882} & \textcolor{gray}{.829} \\
\textcolor{gray}{dino.txt-L/14~\cite{jose2025dinov2}} & \textcolor{gray}{1.038} & \textcolor{gray}{1.145} & \textcolor{gray}{2.005} & \textcolor{gray}{2.231} & \textcolor{gray}{2.321} & \textcolor{gray}{2.930} & \textcolor{gray}{1.945} & \textcolor{gray}{1.204} & \textcolor{gray}{1.226} \\
\midrule
\ours{} (\stageone{})                                & 1.031 & 1.082 & 1.392 & 1.901 & 1.810 & 2.097 & 1.552 & .992 & 1.000 \\
\rowcolor{oursrow}
\textbf{\ours{}}                                     & \textbf{.891} & .932 & 1.128 & \textbf{1.558} & 1.459 & \textbf{1.711} & \textbf{1.280} & \textbf{.820} & \textbf{.829} \\
\bottomrule
\end{tabular}}
\caption{\textbf{LIBERO state probe in centimeters.}
Position error of the objects on held-out episodes, for the current frame (readout) and 1 and 3\,s ahead given the robot actions.
Over the six cells: mean error, mean of the median errors, and mean of the errors relative to \ours{} (\stageone{}), as reported in \cref{tab:main}.
Marks as in \cref{tab:main}.}
\label{tab:libero}
\end{table*}

%% file: tab/libero_suppl.tex
\begin{table*}[t]
\centering
\small
\setlength{\tabcolsep}{5pt}
\fitwidth{\textwidth}{%
\begin{tabular}{l cc cccc}
\toprule
 & orientation & predicates & \multicolumn{4}{c}{position error of the objects that moved (cm\,$\downarrow$)} \\
\cmidrule(lr){4-7}
Encoder & error (deg\,$\downarrow$) & macro-F1\,$\uparrow$ & Object +1\,s & Object +3\,s & Spatial +1\,s & Spatial +3\,s \\
\midrule
\multicolumn{7}{l}{\emph{Spatial \& self-supervised encoders}} \\
TIPSv2-B/14~\cite{cao2026tipsv2}                     & 1.79 & .871 & 3.53 & 3.87 & 2.94 & 3.10 \\
DINOv2-B/14~\cite{oquab2023dinov2}                   & 1.71 & .889 & 3.46 & 5.04 & 3.15 & 3.31 \\
PE-Spatial-B/16~\cite{bolya2026perception}           & 1.84 & .866 & 3.38 & 4.27 & 2.96 & 3.54 \\
DINOv3-B/16~\cite{simeoni2025dinov3}                 & 1.77 & \textbf{.894} & 3.33 & \textbf{3.24} & 2.93 & \textbf{3.04} \\
C-RADIOv3-B~\cite{heinrich2025radiov2}               & 1.75 & .891 & \textbf{3.18} & 3.80 & \textbf{2.82} & 3.19 \\
\midrule
\multicolumn{7}{l}{\emph{Image-text encoders}} \\
CLIP-B/16~\cite{radford2021learning}                 & 1.78 & .849 & 3.71 & 5.32 & 3.54 & 4.91 \\
SigLIP~2-B/16~\cite{tschannen2025siglip}             & 1.81 & .828 & 4.15 & 4.42 & 3.47 & 3.95 \\
PE-Core-B/16~\cite{bolya2026perception}              & 1.78 & .841 & 3.43 & 5.90 & 3.55 & 3.98 \\
OpenVision-B/16~\cite{li2025openvision}              & 1.77 & .853 & 3.65 & 4.82 & 3.17 & 3.40 \\
\midrule
\multicolumn{7}{l}{\emph{Pixel \& latent reconstruction encoders}} \\
MAE-B/16~\cite{he2022masked}                         & 1.78 & .830 & 3.30 & 4.69 & 3.18 & 3.80 \\
VC-1-B~\cite{majumdar2023we}                        & 1.72 & .868 & 3.45 & 3.55 & 3.47 & 3.45 \\
OpenVision~3-B~\cite{zhang2026openvision}            & 1.77 & .847 & 3.29 & 3.99 & 3.50 & 4.00 \\
\midrule
\multicolumn{7}{l}{\emph{Larger encoders, ViT-L (304M to 330M parameters)}} \\
\textcolor{gray}{OpenVision~2-L/14~\cite{liu2026openvision}} & \textcolor{gray}{1.73} & \textcolor{gray}{.900} & \textcolor{gray}{3.24} & \textcolor{gray}{3.60} & \textcolor{gray}{3.11} & \textcolor{gray}{3.44} \\
\textcolor{gray}{V-JEPA~2-L/16~\cite{assran2025v}}   & \textcolor{gray}{1.68} & \textcolor{gray}{.897} & \textcolor{gray}{2.93} & \textcolor{gray}{3.89} & \textcolor{gray}{2.51} & \textcolor{gray}{2.83} \\
\textcolor{gray}{dino.txt-L/14~\cite{jose2025dinov2}} & \textcolor{gray}{1.84} & \textcolor{gray}{.841} & \textcolor{gray}{3.70} & \textcolor{gray}{5.42} & \textcolor{gray}{4.09} & \textcolor{gray}{5.46} \\
\midrule
\ours{} (\stageone{})                                & 1.79 & .863 & 3.52 & 4.14 & 3.29 & 3.82 \\
\rowcolor{oursrow}
\textbf{\ours{}}                                     & \textbf{1.69} & .874 & 3.20 & 3.44 & 3.08 & 3.17 \\
\bottomrule
\end{tabular}}
\caption{\textbf{Further outputs of the state probe.}
Orientation error and macro-F1 of the binary predicates, as means over the six cells, and the position error of the objects that moved between the current and the target frame.
Marks as in \cref{tab:main}.}
\label{tab:libero_suppl}
\end{table*}

%% file: tab/libero_r256.tex
\begin{table}[t]
\centering
\footnotesize
\setlength{\tabcolsep}{2.2pt}
\begin{tabular}{l cccc c c}
\toprule
 & \multicolumn{5}{c}{Frames rendered at 256 px} & 128 px \\
\cmidrule(lr){2-6} \cmidrule(lr){7-7}
 & read- & & & & mean & \\
Encoder & out & +1\,s & +3\,s & mean & (cm) & mean \\
\midrule
TIPSv2-B/14 & .875 & .896 & 1.047 & .939 & 1.25 & .963 \\
DINOv2-B/14 & .863 & .895 & 1.008 & .922 & 1.22 & .942 \\
PE-Spatial-B/16 & .986 & .901 & .923 & .937 & 1.24 & .936 \\
DINOv3-B/16 & .880 & .901 & .806 & .862 & 1.15 & .870 \\
C-RADIOv3-B & .881 & .831 & .973 & .895 & 1.19 & .864 \\
\midrule
CLIP-B/16 & .980 & 1.056 & 1.201 & 1.079 & 1.43 & 1.194 \\
SigLIP~2-B/16 & 1.102 & 1.069 & 1.028 & 1.066 & 1.42 & 1.160 \\
PE-Core-B/16 & 1.045 & 1.003 & 1.185 & 1.078 & 1.44 & 1.170 \\
OpenVision-B/16 & 1.050 & 1.037 & .990 & 1.026 & 1.37 & 1.027 \\
\midrule
MAE-B/16 & .868 & .942 & .944 & .918 & 1.24 & .990 \\
VC-1-B & .978 & .918 & .953 & .950 & 1.27 & .943 \\
OpenVision~3-B & .887 & .926 & .947 & .920 & 1.24 & 1.011 \\
\midrule
\textcolor{gray}{OpenVision~2-L/14} & \textcolor{gray}{.962} & \textcolor{gray}{.903} & \textcolor{gray}{.912} & \textcolor{gray}{.926} & \textcolor{gray}{1.25} & \textcolor{gray}{.945} \\
\textcolor{gray}{V-JEPA~2-L/16} & \textcolor{gray}{.743} & \textcolor{gray}{.885} & \textcolor{gray}{.940} & \textcolor{gray}{.856} & \textcolor{gray}{1.14} & \textcolor{gray}{.829} \\
\textcolor{gray}{dino.txt-L/14} & \textcolor{gray}{1.047} & \textcolor{gray}{1.038} & \textcolor{gray}{1.051} & \textcolor{gray}{1.045} & \textcolor{gray}{1.40} & \textcolor{gray}{1.226} \\
\midrule
\ours{} (\stageone{}) & 1.000 & 1.000 & 1.000 & 1.000 & 1.34 & 1.000 \\
\ours{} & .868 & .856 & .811 & \textbf{.845} & 1.13 & .829 \\
\bottomrule
\end{tabular}
\caption{\textbf{State probe on frames rendered at $256\times256$ resolution} from the stored states of the episodes of \cref{tab:main}, with the same splits, targets, and probe.
Relative error with \ours{} (\stageone{}) on the same frames as 1.000; last column: the 128-px mean of \cref{tab:main}.
Bold: lowest ViT-B mean; \textcolor{gray}{gray}: ViT-L.}
\label{tab:libero_r256}
\end{table}

%% file: tab/layers.tex
\begin{table}[t]
\centering
\small
\setlength{\tabcolsep}{3.1pt}
\begin{tabular}{l c c cccc}
\toprule
 & & Validation & \multicolumn{4}{c}{Relative error\,$\downarrow$} \\
\cmidrule(lr){4-7}
 & & loss & read- & & & \\
Encoder & Block & (mean) & out & +1\,s & +3\,s & mean \\
\midrule
SigLIP~2-B/16 & 6 & \textbf{.046} & 1.194 & .970 & 1.277 & \textbf{1.147} \\
 & 8 & .052 & 1.165 & 1.106 & 1.249 & 1.173 \\
 & 10 & .056 & 1.260 & 1.169 & 1.368 & 1.266 \\
 & 12 & .059 & 1.149 & 1.202 & 1.130 & 1.160 \\
\midrule
PE-Core-B/16 & 6 & \textbf{.045} & 1.039 & .861 & 1.126 & \textbf{1.009} \\
 & 8 & .048 & .943 & .909 & 1.104 & .985 \\
 & 10 & .046 & .908 & .975 & 1.081 & .988 \\
 & 12 & .054 & 1.118 & 1.099 & 1.292 & 1.170 \\
\midrule
\ours{} & 12 & .039 & .842 & .834 & .813 & .829 \\
\bottomrule
\end{tabular}
\caption{\textbf{Intermediate layers on the LIBERO state probe.}
Relative error as in \cref{tab:main} for blocks 6, 8, and 10 and for the last block, 12, which \cref{tab:main} uses.
Bold: the block with the lowest mean validation loss, a selection rule fixed before the runs; \ours{} uses its last block.}
\label{tab:layers}
\end{table}

%% file: tab/policy_detail.tex
\begin{table*}[t]
\centering
\small
\setlength{\tabcolsep}{4.2pt}
\fitwidth{\textwidth}{%
\begin{tabular}{l ccc ccc ccc ccc c}
\toprule
 & \multicolumn{3}{c}{LIBERO-Long} & \multicolumn{3}{c}{LIBERO-Goal} & \multicolumn{3}{c}{LIBERO-Spatial} & \multicolumn{3}{c}{LIBERO-Object} & All four \\
\cmidrule(lr){2-4} \cmidrule(lr){5-7} \cmidrule(lr){8-10} \cmidrule(lr){11-13} \cmidrule(lr){14-14}
Encoder & success & worst task & tasks $\geq$ 80 & success & worst task & tasks $\geq$ 80 & success & worst task & tasks $\geq$ 80 & success & worst task & tasks $\geq$ 80 & mean \\
\midrule
CLIP-B/16                                      & 56.3 {\scriptsize [43, 69]} & 23 & 1 & 74.1 {\scriptsize [63, 84]} & 37 & 4 & 76.5 {\scriptsize [70, 82]} & 56 & 5 & 93.2 {\scriptsize [89, 97]} & 79 & 9 & 75.0 \\
\quad with its text tower                      & 61.5 {\scriptsize [50, 73]} & 31 & 1 & 84.3 {\scriptsize [74, 93]} & 53 & 7 & 79.1 {\scriptsize [71, 87]} & 47 & 6 & 91.9 {\scriptsize [89, 95]} & 87 & \textbf{10} & 79.2 \\
SigLIP~2-B/16                                  & 72.3 {\scriptsize [55, 87]} & 29 & \textbf{6} & 78.8 {\scriptsize [68, 88]} & 47 & 5 & 84.1 {\scriptsize [75, 92]} & 55 & \textbf{8} & 90.3 {\scriptsize [84, 95]} & 67 & 9 & 81.4 \\
\quad with its text tower                      & 73.3 {\scriptsize [63, 83]} & 46 & 3 & \textbf{90.2} {\scriptsize [83, 96]} & 69 & 8 & 84.1 {\scriptsize [77, 91]} & 68 & 6 & 91.7 {\scriptsize [87, 96]} & 77 & 9 & 84.8 \\
DINOv2-B/14                                    & 66.5 {\scriptsize [54, 78]} & 29 & 3 & 78.6 {\scriptsize [71, 87]} & 63 & 4 & 80.5 {\scriptsize [74, 87]} & 62 & 4 & 88.1 {\scriptsize [82, 93]} & 69 & 9 & 78.4 \\
DINOv3-B/16                                    & 68.4 {\scriptsize [57, 78]} & 40 & 4 & 72.7 {\scriptsize [63, 82]} & 51 & 3 & 80.7 {\scriptsize [73, 87]} & 52 & 7 & \textbf{94.4} {\scriptsize [92, 97]} & \textbf{89} & \textbf{10} & 79.0 \\
C-RADIOv3-B                                    & 66.0 {\scriptsize [53, 78]} & 39 & 5 & 79.9 {\scriptsize [70, 89]} & 53 & 6 & 74.9 {\scriptsize [67, 83]} & 58 & 4 & 92.3 {\scriptsize [88, 96]} & 77 & 9 & 78.2 \\
\quad with its text tower                      & 73.4 {\scriptsize [61, 84]} & 36 & 4 & 89.5 {\scriptsize [83, 94]} & 67 & \textbf{9} & 82.9 {\scriptsize [76, 89]} & 63 & 7 & 81.4 {\scriptsize [66, 93]} & 24 & 7 & 81.8 \\
\midrule
\ours{} (\stageone{})                          & 68.9 {\scriptsize [57, 80]} & 33 & 2 & 84.3 {\scriptsize [76, 91]} & 61 & 7 & 76.8 {\scriptsize [65, 86]} & 32 & 5 & 94.3 {\scriptsize [92, 97]} & 88 & \textbf{10} & 81.1 \\
\textbf{\ours{}}                               & 69.7 {\scriptsize [59, 80]} & 44 & 3 & 85.1 {\scriptsize [79, 91]} & \textbf{71} & 6 & \textbf{86.5} {\scriptsize [81, 91]} & \textbf{73} & 7 & 92.2 {\scriptsize [88, 96]} & 79 & 9 & 83.4 \\
\quad with its text tower                      & \textbf{78.5} {\scriptsize [68, 88]} & \textbf{51} & 5 & 87.4 {\scriptsize [79, 94]} & 57 & 8 & 82.6 {\scriptsize [74, 90]} & 55 & 5 & 93.7 {\scriptsize [90, 97]} & 84 & \textbf{10} & \textbf{85.5} \\
\bottomrule
\end{tabular}}
\caption{\textbf{Policy learning on LIBERO in detail, for the policies of \cref{tab:policy}.}
Success in \% with a 95\% bootstrap interval over tasks and initial states; worst task: the lowest of the 10 tasks; tasks $\geq$ 80: tasks with at least 80\% success.
Indented rows: instructions embedded by the encoder's own text tower instead of BERT. Bold: best of each column.}
\label{tab:policy_detail}
\end{table*}

%% file: tab/pertask_suppl.tex
\begin{table*}[p]
\centering
\footnotesize
\setlength{\tabcolsep}{3pt}
\begin{tabular}{l ccccc cc}
\toprule
Task & CLIP & SigLIP~2 & DINOv2 & DINOv3 & C-RADIOv3 & \shortstack{\ours{}\\ (\stageone{})} & \ours{} \\
\midrule
\multicolumn{8}{l}{\emph{LIBERO-Long}} \\
alphabet soup and tomato sauce in the basket                       & 48 & \textbf{71} & 59 & 65 & 39 & 62 & \textbf{71} \\
cream cheese box and butter in the basket                          & 51 & \textbf{91} & 71 & 86 & 65 & 68 & 71 \\
turn on the stove, moka pot on it                                  & 79 & \textbf{97} & 85 & 81 & 85 & 95 & 79 \\
black bowl in the bottom drawer, close it                          & 90 & \textbf{98} & 94 & 87 & 93 & 96 & 93 \\
white mug on the left plate, yellow-white mug on the right plate   & 42 & 50 & 48 & 51 & 49 & \textbf{61} & 55 \\
book in the back compartment of the caddy                          & 58 & \textbf{91} & 68 & 75 & 83 & 73 & 68 \\
white mug on the plate, pudding right of the plate                 & 29 & 29 & \textbf{52} & 40 & 43 & 33 & 47 \\
alphabet soup and cream cheese box in the basket                   & 67 & \textbf{85} & 79 & 82 & 83 & 69 & 81 \\
both moka pots on the stove                                        & 23 & 30 & 29 & 47 & 39 & \textbf{53} & 44 \\
yellow-white mug in the microwave, close it                        & 75 & 82 & 81 & 71 & 81 & 78 & \textbf{89} \\
\emph{mean} & 56.3 & 72.3 & 66.5 & 68.4 & 66.0 & 68.9 & 69.7 \\
\midrule
\multicolumn{8}{l}{\emph{LIBERO-Goal}} \\
open the middle drawer of the cabinet                              & 93 & 96 & 92 & 85 & \textbf{97} & 92 & 95 \\
put the bowl on the stove                                          & 81 & 91 & 91 & 79 & 94 & \textbf{98} & 96 \\
put the wine bottle on top of the cabinet                          & \textbf{84} & 83 & 63 & 59 & 53 & 83 & 76 \\
open the top drawer and put the bowl inside                        & 55 & 71 & 63 & 61 & 68 & \textbf{81} & 77 \\
put the bowl on top of the cabinet                                 & 37 & 47 & \textbf{81} & 61 & 80 & 74 & 71 \\
push the plate to the front of the stove                           & 77 & 89 & 69 & 73 & 83 & \textbf{93} & 86 \\
put the cream cheese in the bowl                                   & 73 & 62 & 67 & 51 & 60 & 72 & \textbf{78} \\
turn on the stove                                                  & 97 & 98 & \textbf{100} & \textbf{100} & 96 & 96 & 99 \\
put the bowl on the plate                                          & 65 & 74 & 79 & 89 & \textbf{93} & \textbf{93} & 92 \\
put the wine bottle on the rack                                    & 79 & 77 & 79 & 69 & 75 & 61 & \textbf{81} \\
\emph{mean} & 74.1 & 78.8 & 78.6 & 72.7 & 79.9 & 84.3 & 85.1 \\
\midrule
\multicolumn{8}{l}{\emph{LIBERO-Spatial: pick up the black bowl \ldots{} and place it on the plate}} \\
between the plate and the ramekin                                  & 80 & 87 & 87 & 87 & 85 & 87 & \textbf{93} \\
next to the ramekin                                                & 73 & \textbf{86} & 76 & 81 & 81 & 81 & 79 \\
from the table center                                              & 82 & \textbf{96} & 92 & 95 & 69 & 86 & 94 \\
on the cookie box                                                  & 77 & \textbf{98} & 87 & 89 & 58 & 89 & 94 \\
in the top drawer of the wooden cabinet                            & 79 & 80 & 62 & 81 & 87 & 74 & \textbf{89} \\
on the ramekin                                                     & 56 & 55 & 73 & 52 & 70 & 32 & \textbf{79} \\
next to the cookie box                                             & 87 & \textbf{98} & 96 & 91 & 96 & 97 & \textbf{98} \\
on the stove                                                       & 81 & \textbf{91} & 78 & 69 & 72 & 75 & 73 \\
next to the plate                                                  & 65 & \textbf{85} & 75 & 83 & 61 & 70 & 81 \\
on the wooden cabinet                                              & 83 & 65 & 78 & 79 & 69 & 77 & \textbf{85} \\
\emph{mean} & 76.5 & 84.1 & 80.5 & 80.7 & 74.9 & 76.8 & 86.5 \\
\midrule
\multicolumn{8}{l}{\emph{LIBERO-Object: pick up the \ldots{} and place it in the basket}} \\
alphabet soup                                                      & \textbf{97} & 67 & 95 & 91 & 93 & 91 & \textbf{97} \\
cream cheese                                                       & 79 & \textbf{95} & 69 & 93 & 93 & \textbf{95} & 90 \\
salad dressing                                                     & 97 & 99 & 94 & \textbf{100} & 97 & 99 & 97 \\
bbq sauce                                                          & 89 & \textbf{99} & 97 & 95 & 94 & 97 & 86 \\
ketchup                                                            & 92 & 83 & 87 & \textbf{97} & 85 & 94 & 93 \\
tomato sauce                                                       & 93 & 96 & 86 & 89 & \textbf{97} & 89 & 91 \\
butter                                                             & \textbf{100} & 93 & 89 & 91 & 97 & 88 & 96 \\
milk                                                               & 91 & 85 & 83 & 93 & 77 & \textbf{95} & 79 \\
chocolate pudding                                                  & \textbf{99} & 93 & \textbf{99} & \textbf{99} & 92 & 98 & 97 \\
orange juice                                                       & 93 & 92 & 81 & 96 & \textbf{98} & 96 & 97 \\
\emph{mean} & 93.2 & 90.3 & 88.1 & 94.4 & 92.3 & 94.3 & 92.2 \\
\bottomrule
\end{tabular}
\caption{\textbf{Success of every task} in \% for the policies of \cref{tab:policy} with BERT, each value the mean of three evaluations (150 rollouts per task), tasks in benchmark order; the mean of a suite is the value of \cref{tab:policy}. Bold: best of a row.}
\label{tab:pertask}
\end{table*}

%% file: tab/policy_hparams.tex
\begin{table*}[tp]
\centering
\small
\setlength{\tabcolsep}{4pt}
\begin{tabular}{ll}
\toprule
\multicolumn{2}{l}{\emph{Observation}} \\
cameras & agent view and eye-in-hand, $128 \times 128$, flipped upright \\
encoder input & bilinear resize to $224 \times 224$, normalization of the encoder \\
encoder precision & bfloat16 autocast, tokens cast to 32 bit \\
tokens per camera & 196 patch tokens of the last block (DINOv2: 256) \\
pooling & per camera, one query from the task embedding, 4 heads, output 64 \\
language & BERT-base (cased), pooled embedding, frozen \\
proprioception & joint positions (7) and gripper state (2) \\
\midrule
\multicolumn{2}{l}{\emph{Policy}} \\
architecture & transformer policy of LIBERO~\cite{liu2023libero} \\
transformer & 4 layers, 6 heads, width 64, MLP 256, dropout 0.1 \\
context & 10 steps \\
action head & Gaussian mixture with 5 modes \\
trainable parameters & 8.8M, the same for every encoder \\
\midrule
\multicolumn{2}{l}{\emph{Training}} \\
data & 50 demonstrations per task, 10 tasks per suite \\
optimizer & AdamW, learning rate $10^{-4}$, weight decay $10^{-4}$ \\
schedule & cosine decay to $10^{-5}$, no warm-up \\
batch & 32 windows of 10 steps \\
epochs & 25 on LIBERO-Long, 50 on the other suites \\
augmentation & color jitter ($p = 0.9$), random shift of up to 4 pixels \\
gradient clipping & norm 100 \\
\midrule
\multicolumn{2}{l}{\emph{Evaluation}} \\
checkpoint & last epoch \\
rollouts & 50 per task, one for each fixed initial state of LIBERO \\
episode & at most 600 steps, success when the goal predicates hold \\
repetitions & three evaluations, differing only in the sampled actions \\
\bottomrule
\end{tabular}
\caption{\textbf{Settings of policy learning}; within a suite, every encoder is trained and evaluated with matched settings.}
\label{tab:policy_hparams}
\end{table*}

%% file: tab/rope_suppl.tex
\begin{table}[t]
\centering
\small
\setlength{\tabcolsep}{4.8pt}
\begin{tabular}{l cc}
\toprule
 & \multicolumn{2}{c}{Chamfer\,$\downarrow$, 50 instances} \\
\cmidrule(lr){2-3}
Encoder & mean & median \\
\midrule
DINOv2-B/14                          & .701 & .446 \\
DINOv3-B/16                          & .681 & .406 \\
C-RADIOv3-B                          & .558 & .289 \\
SigLIP~2-B/16                        & .848 & .511 \\
\midrule
\ours{} (\stageone{})                & .943 & .514 \\
\ours{}                              & .750 & .448 \\
\midrule
DINOv2-S/14, recipe of DINO-WM       & .547 & .309 \\
\bottomrule
\end{tabular}
\caption{\textbf{Planning on Rope} with the planner of \cref{tab:planning}.
The first six rows use the world-model recipe of \cref{tab:planning}; the last row uses the predictor and the schedule of DINO-WM with its encoder, DINOv2-S/14, as a check of the pipeline.}
\label{tab:rope}
\end{table}

%% file: tab/reacher_suppl.tex
\begin{table}[t]
\centering
\small
\setlength{\tabcolsep}{4pt}
\fitwidth{\columnwidth}{%
\begin{tabular}{l ccc c}
\toprule
 & \multicolumn{3}{c}{seed of the planner} & \\
\cmidrule(lr){2-4}
Encoder & 42 & 43 & 44 & mean \\
\midrule
DINOv2-B/14                                    & 72 & 80 & 78 & 76.7 \\
DINOv3-B/16                                    & 90 & 90 & 78 & 86.0 \\
C-RADIOv3-B                                    & 82 & 86 & 78 & 82.0 \\
SigLIP~2-B/16                                  & 68 & 78 & 76 & 74.0 \\
\midrule
\ours{} (\stageone{})                          & 70 & 78 & 80 & 76.0 \\
\ours{}                                        & 82 & 80 & 74 & 78.7 \\
\bottomrule
\end{tabular}}
\caption{\textbf{Reacher success in \% for every planner seed}, with the same world model and the same 50 episodes; the mean is the value of \cref{tab:planning}.}
\label{tab:reacher_seeds}
\end{table}

%% file: tab/wmprobe_suppl.tex
\begin{table*}[t]
\centering
\footnotesize
\setlength{\tabcolsep}{4pt}
\renewcommand{\arraystretch}{1.1}
\begin{tabular}{@{}>{\raggedright\arraybackslash}p{0.14\textwidth} *{5}{>{\raggedright\arraybackslash}p{0.1555\textwidth}}@{}}
\toprule
\multicolumn{6}{@{}l}{\emph{Shared by all encoders and tasks}} \\
Input & \multicolumn{5}{>{\raggedright\arraybackslash}p{0.8425\textwidth}@{}}{the 196 tokens of one frame with their 768 visual channels, standardized as for the world model; real tokens in training, real or predicted tokens in evaluation; no proprioceptive or action channels and no history frames} \\
Architecture & \multicolumn{5}{>{\raggedright\arraybackslash}p{0.8425\textwidth}@{}}{linear projection to width 256 with learned token embeddings; 8 learned queries; 2 cross-attention blocks with 8 heads and an MLP of width 1{,}024; layer normalization and a linear layer on the concatenated queries; 1.8M parameters, 3.4M for 256 points} \\
Data & \multicolumn{5}{>{\raggedright\arraybackslash}p{0.8425\textwidth}@{}}{real tokens of 20{,}000 frames of the training data; 2{,}000 other frames of the training data for model selection; evaluation on held-out trajectories} \\
Optimization & \multicolumn{5}{>{\raggedright\arraybackslash}p{0.8425\textwidth}@{}}{AdamW, learning rate $3\times10^{-4}$, weight decay 0.05, cosine decay to zero without warm-up, no gradient clipping, batch 128, 6{,}000 updates, seed 0; encoder frozen} \\
Selection & \multicolumn{5}{>{\raggedright\arraybackslash}p{0.8425\textwidth}@{}}{lowest validation loss, computed every 500 updates; every selected checkpoint lies at 5{,}000 to 6{,}000 updates} \\
Evaluation & \multicolumn{5}{>{\raggedright\arraybackslash}p{0.8425\textwidth}@{}}{probe frozen and in evaluation mode; one probe per encoder and task reads the real and predicted tokens of every world model of that encoder} \\
\midrule
\emph{Per task} & PushT & Wall & Granular & Rope & Reacher \\
\midrule
World-model history & 3 frames, skip 5 & 1 frame, skip 5 & 1 frame, no skip & 1 frame, no skip & 3 frames, skip 5 \\
Proprioceptive input of the world model & agent position and velocity & agent position & constant & constant & none \\
Probe frames & 8-frame windows, \mbox{skip 5} & 8-frame windows, \mbox{skip 5} & 8-frame windows, \mbox{no skip} & 8-frame windows, \mbox{no skip} & every 5th step of 100-step blocks \\
Probe target & agent and block position, block angle (sine, cosine) & agent position & 256 points $(x, y, z)$ & 256 points $(x, y, z)$ & two joint angles (sine, cosine), fingertip position \\
Training loss & MSE & MSE & Chamfer to a fixed subset of 2{,}048 particles & Chamfer to all 1{,}965 particles & MSE \\
Error in \cref{tab:predacc} & block position, workspace units & agent position, workspace units & Chamfer distance & Chamfer distance & joint angles, degrees \\
\bottomrule
\end{tabular}
\caption{\textbf{State probe of \cref{tab:predacc,tab:predacc_full}}: settings shared by all encoders and tasks, and per task, with the inputs that only the world model receives.
MSE: mean squared error of targets standardized with training statistics; Chamfer: symmetric Chamfer distance between the predicted points and the particles of the true state.}
\label{tab:wmprobe}
\end{table*}

%% file: tab/decode_suppl.tex
\begin{table}[t]
\centering
\footnotesize
\setlength{\tabcolsep}{3.5pt}
{%
\begin{tabular}{l cc}
\toprule
Encoder & $\ell_1$ & LPIPS \\
\midrule
DINOv2-B/14                 & .0065 & .0091 \\
DINOv3-B/16                 & .0065 & .0080 \\
C-RADIOv3-B                 & .0062 & .0074 \\
SigLIP~2-B/16               & .0066 & .0086 \\
\midrule
\ours{} (\stageone{})       & \textbf{.0055} & \textbf{.0069} \\
\ours{}                    & .0061 & .0073 \\
\midrule
SDXL autoencoder alone      & .0044 & .0046 \\
mean training frame         & .0370 & .0866 \\
\bottomrule
\end{tabular}}
\caption{\textbf{Decoding PushT frames from frozen tokens.}
We use the protocol of \cref{tab:main} and evaluate on 1{,}000 frames from validation trajectories.
Lower is better; bold: best encoder in each column.}
\label{tab:decode}
\end{table}

%% file: tab/render.tex
\begin{table}[t]
\centering
\footnotesize
\setlength{\tabcolsep}{4.5pt}
\begin{tabular}{l c c c cc}
\toprule
 & token & decoded real & \multicolumn{3}{c}{rendered prediction} \\
\cmidrule(lr){4-6}
 & cosine & $k = 5$ & $k = 1$ & \multicolumn{2}{c}{$k = 5$} \\
Encoder & & LPIPS & LPIPS & $\ell_1$ & LPIPS \\
\midrule
SigLIP~2-B/16                            & 1.000 & .0081 & .0107 & .0150 & .0308 \\
C-RADIOv3-B                              & 1.000 & .0070 & .0088 & \textbf{.0103} & \textbf{.0176} \\
DINOv3-B/16                              & 1.000 & .0074 & .0096 & .0128 & .0242 \\
\midrule
\ours{} (\stageone{})                    & 1.000 & \textbf{.0066} & .0093 & .0126 & .0248 \\
\ours{}                                  & 1.000 & .0068 & .0090 & .0114 & .0203 \\
\bottomrule
\end{tabular}
\caption{\textbf{Rendered world-model predictions on PushT} (200 windows; lower is better).
Token cosine: world-model vs.\ decoder-training tokens (DINOv2: .80, not rendered).}
\label{tab:render}
\end{table}

%% file: tab/predacc_full_suppl.tex
\begin{table*}[tp]
\centering
\small
\setlength{\tabcolsep}{7pt}
\begin{tabular}{l ccccc cc}
\toprule
 & \multicolumn{5}{c}{predicted, world-model steps ahead} & \multicolumn{2}{c}{references, 5 steps ahead} \\
\cmidrule(lr){2-6} \cmidrule(lr){7-8}
Encoder & 1 & 2 & 3 & 4 & 5 & true tokens & last obs. \\
\multicolumn{8}{l}{\emph{PushT} (block position, workspace units)} \\
DINOv2-B/14                              & 4.29 & 4.47 & 5.16 & 6.08 & 7.16 & 3.24 & 38.27 \\
DINOv3-B/16                              & 3.87 & 4.38 & 5.47 & 7.04 & 8.70 & 2.93 & 37.87 \\
C-RADIOv3-B                              & 3.51 & 3.66 & 4.21 & 4.85 & 5.64 & 2.66 & 38.14 \\
SigLIP~2-B/16                            & 6.17 & 6.87 & 8.21 & 9.97 & 11.96 & 3.52 & 38.02 \\
\ours{} (\stageone{})                    & 5.49 & 5.66 & 6.58 & 7.77 & 9.32 & 3.36 & 38.13 \\
\ours{}                                  & 4.47 & 4.74 & 5.29 & 5.85 & 6.73 & 2.85 & 37.87 \\
\midrule
\multicolumn{8}{l}{\emph{Wall} (agent position, workspace units)} \\
DINOv2-B/14                              & 6.74 & 7.47 & 8.22 & 8.94 & 9.69 & 9.74 & 10.73 \\
DINOv3-B/16                              & 6.73 & 7.48 & 8.26 & 9.00 & 9.78 & 9.79 & 10.72 \\
C-RADIOv3-B                              & 6.73 & 7.45 & 8.24 & 8.98 & 9.75 & 9.75 & 10.71 \\
SigLIP~2-B/16                            & 6.90 & 7.57 & 8.42 & 9.16 & 9.91 & 9.77 & 10.75 \\
\ours{} (\stageone{})                    & 6.71 & 7.46 & 8.28 & 9.03 & 9.81 & 9.75 & 10.73 \\
\ours{}                                  & 6.71 & 7.43 & 8.21 & 8.94 & 9.69 & 9.76 & 10.71 \\
\midrule
\multicolumn{8}{l}{\emph{Granular} (Chamfer distance)} \\
DINOv2-B/14                              & .215 & .225 & .233 & .240 & .246 & .225 & .599 \\
DINOv3-B/16                              & .192 & .202 & .210 & .217 & .222 & .199 & .602 \\
C-RADIOv3-B                              & .192 & .201 & .208 & .214 & .218 & .200 & .600 \\
SigLIP~2-B/16                            & .228 & .243 & .253 & .261 & .268 & .231 & .603 \\
\ours{} (\stageone{})                    & .229 & .244 & .255 & .265 & .272 & .223 & .595 \\
\ours{}                                  & .203 & .214 & .223 & .231 & .236 & .208 & .597 \\
\midrule
\multicolumn{8}{l}{\emph{Rope} (Chamfer distance)} \\
DINOv2-B/14                              & .142 & .204 & .265 & .335 & .410 & .055 & 1.674 \\
DINOv3-B/16                              & .133 & .195 & .259 & .330 & .407 & .052 & 1.680 \\
C-RADIOv3-B                              & .126 & .179 & .240 & .299 & .372 & .051 & 1.678 \\
SigLIP~2-B/16                            & .178 & .253 & .320 & .399 & .469 & .066 & 1.669 \\
\ours{} (\stageone{})                    & .161 & .228 & .295 & .360 & .424 & .056 & 1.674 \\
\ours{}                                  & .135 & .195 & .248 & .313 & .377 & .053 & 1.678 \\
\midrule
\multicolumn{8}{l}{\emph{Reacher} (joint angles, degrees)} \\
DINOv2-B/14                              & .90 & .99 & 1.08 & 1.14 & 1.21 & .93 & 25.23 \\
DINOv3-B/16                              & .70 & .81 & .88 & .95 & 1.03 & .75 & 25.23 \\
C-RADIOv3-B                              & .81 & .89 & .95 & 1.01 & 1.06 & .80 & 25.21 \\
SigLIP~2-B/16                            & 1.10 & 1.39 & 1.62 & 1.84 & 2.02 & 1.00 & 25.24 \\
\ours{} (\stageone{})                    & .90 & 1.10 & 1.20 & 1.30 & 1.42 & .89 & 25.22 \\
\ours{}                                  & .83 & .93 & 1.02 & 1.09 & 1.18 & .79 & 25.22 \\
\bottomrule
\end{tabular}
\caption{\textbf{State readout from predicted tokens at every step.}
We evaluate the state probe of \cref{tab:predacc} on tokens predicted 1 to 5 world-model steps ahead, alongside two references scored against the true state at step 5: the true tokens at step 5 and the last observed tokens (lower is better).}
\label{tab:predacc_full}
\end{table*}

%% file: tab/wmdiag.tex
\begin{table*}[tp]
\centering
\small
\setlength{\tabcolsep}{3.3pt}
\begin{tabular}{l ccccc}
\toprule
 & true future & one-step prediction, & five-step & rollout with recorded & last observation \\
Encoder & tokens & true history & open-loop rollout & proprioception & repeated \\
\midrule
\ours{}                        & 3.1 & 3.6 & 6.6 & 6.7 & 42.0 \\
\ours{} (\stageone{})          & 3.8 & 4.3 & 9.9 & 9.2 & 42.3 \\
C-RADIOv3-B                    & \textbf{2.4} & \textbf{2.8} & \textbf{5.8} & \textbf{5.1} & 41.9 \\
DINOv3-B/16                    & 3.0 & 3.2 & 9.1 & 8.4 & 41.7 \\
SigLIP~2-B/16                  & 3.7 & 5.2 & 14.1 & 12.1 & 42.0 \\
\midrule
\ours{} $-$ C-RADIOv3 & $+0.68$ {\scriptsize [$+0.26$, $+1.22$]} & $+0.73$ {\scriptsize [$+0.38$, $+1.09$]} & $+0.76$ {\scriptsize [$-0.04$, $+1.49$]} & $+1.60$ {\scriptsize [$+0.42$, $+2.97$]} & $+0.13$ {\scriptsize [$-0.60$, $+0.80$]} \\
\bottomrule
\end{tabular}
\caption{\textbf{PushT block error read from true, one-step, and rolled-out tokens 5 world-model steps ahead}, in workspace units, lower is better, on the 200 windows of \cref{tab:render} with the world models of \cref{tab:planning}.
One-step: every step predicted from the true history. Last row: paired difference with a 95\% bootstrap interval over the 21 trajectories. Bold: lowest of a column; conditions: \cref{sec:suppl_wmdiag}.}
\label{tab:wmdiag}
\end{table*}

%% file: tab/recon_large.tex
\begin{table}[t]
\centering
\small
\setlength{\tabcolsep}{5pt}
\begin{tabular}{l cc cc}
\toprule
 & \multicolumn{2}{c}{Standard decoder} & \multicolumn{2}{c}{Larger decoder} \\
\cmidrule(lr){2-3} \cmidrule(lr){4-5}
Encoder & $\ell_1$ & LPIPS & $\ell_1$ & LPIPS \\
\midrule
\ours{} & .120 & .372 & .107 & .302 \\
C-RADIOv3-B & .161 & .359 & .141 & .311 \\
DINOv2-B/14 & .248 & .535 & .218 & .423 \\
SigLIP~2-B/16 & .264 & .541 & .220 & .404 \\
\bottomrule
\end{tabular}
\caption{\textbf{Reconstruction with a stronger decoding and training configuration.}
Standard: the protocol of \cref{tab:main}; larger: a wider and deeper decoder trained longer on more images; both on the same 1{,}000 held-out images.
\textbf{Every encoder improves; \ours{} keeps the lowest $\ell_1$ and overtakes C-RADIOv3 on LPIPS.}}
\label{tab:recon_large}
\end{table}

%% file: tab/video.tex
\begin{table}[t]
\centering
\small
\setlength{\tabcolsep}{6pt}
\fitwidth{\linewidth}{%
\begin{tabular}{l cc}
\toprule
 & DAVIS & SSv2 \\
Encoder & $\mathcal{J}\&\mathcal{F}\,\uparrow$ & top-1\,$\uparrow$ \\
\midrule
\multicolumn{3}{l}{\emph{Spatial \& self-supervised encoders}} \\
TIPSv2-B/14~\cite{cao2026tipsv2}                     & .652 & 57.8 \\
DINOv2-B/14~\cite{oquab2023dinov2}                   & .661 & \textbf{63.7} \\
DINOv3-B/16~\cite{simeoni2025dinov3}                 & \textbf{.699} & 61.8 \\
C-RADIOv3-B~\cite{heinrich2025radiov2}               & .659 & 62.1 \\
\midrule
\multicolumn{3}{l}{\emph{Image-text encoders}} \\
CLIP-B/16~\cite{radford2021learning}                 & .574 & 43.3 \\
SigLIP~2-B/16~\cite{tschannen2025siglip}             & .503 & 49.4 \\
PE-Core-B/16~\cite{bolya2026perception}              & .392 & 42.2 \\
\midrule
\multicolumn{3}{l}{\emph{Larger encoders, ViT-L (304M parameters)}} \\
\textcolor{gray}{V-JEPA~2-L/16~\cite{assran2025v}}   & \textcolor{gray}{.526} & \textcolor{gray}{70.3} \\
\midrule
\ours{} (\stageone{})                                & .456 & 56.4 \\
\rowcolor{oursrow}
\textbf{\ours{}}                                     & .533 & 55.6 \\
\bottomrule
\end{tabular}}
\caption{\textbf{Temporal transfer}: DAVIS-2017 video object segmentation by label propagation ($\mathcal{J}\&\mathcal{F}$) and Something-Something~v2 action recognition with an attentive probe (top-1), protocol in \cref{sec:suppl_video}.
SSv2 is in the pretraining data of V-JEPA~2. Encoders without a video result are not listed. Marks as in \cref{tab:main}.}
\label{tab:video}
\end{table}

%% file: tab/stage2_full.tex
\begin{table}[t]
\centering
\footnotesize
\setlength{\tabcolsep}{4.5pt}
\begin{tabular}{l c ccc}
\toprule
 & State probe & \multicolumn{3}{c}{Language alignment\,$\uparrow$} \\
\cmidrule(lr){2-2} \cmidrule(lr){3-5}
 & LIBERO & IN-1k & Flickr & COCO \\
Encoder & err.\,$\downarrow$ & 0-shot & I$\to$T & I$\to$T \\
\midrule
\ours{} (\stageone{}) & 1.000 & 74.5 & 93.4 & 70.6 \\
\ours{}               & .829  & 73.4 & 93.2 & 70.2 \\
\quad\emph{change}    & $-.171$ & $-1.1$ & $-0.2$ & $-0.4$ \\
\midrule
SigLIP~2-B/16         & 1.160 & 77.6 & 91.7 & 67.9 \\
\quad + \stagetwo{}   & .893  & 73.3 & 84.2 & 56.5 \\
\quad\emph{change}    & $-.267$ & $-4.3$ & $-7.5$ & $-11.4$ \\
\midrule
CLIP-B/16             & 1.194 & 68.4 & 81.9 & 52.5 \\
\quad + \stagetwo{}   & .898  & 58.2 & 71.0 & 40.7 \\
\quad\emph{change}    & $-.296$ & $-10.2$ & $-10.9$ & $-11.8$ \\
\bottomrule
\end{tabular}
\caption{\textbf{Numerical results for \cref{fig:refine}}: state-probe error and language alignment for each encoder before and after \stagetwo{}, and the change (after minus before), under the same budget of 5M sampled images.}
\label{tab:stage2_full}
\end{table}